\pdfoutput=1

\documentclass[11pt]{article}

\usepackage[final]{acl}

\usepackage{times}
\usepackage{latexsym}

\usepackage[T1]{fontenc}

\usepackage[utf8]{inputenc}

\usepackage{microtype}

\usepackage{inconsolata}

\usepackage{graphicx}

\usepackage{booktabs}
\usepackage{amsmath}
\usepackage{amssymb}
\usepackage{todonotes}
\usepackage{multirow}
\usepackage{hyperref}
\usepackage{arydshln}

\usepackage{subcaption} 
\usepackage{array}   
\usepackage{calc}     
\usepackage{multirow} 
\usepackage{hhline}   

\usepackage{pifont}
\newcommand{\xmark}{\text{\ding{55}}}
\newcommand{\cmark}{\text{\ding{51}}}

\usepackage{soul} 

\definecolor{lightblue}{RGB}{236,244,255}

\newcommand{\highlight}[1]{\sethlcolor{lightblue}\hl{#1}}

\title{Improving Language and Modality Transfer in Translation by Character-level Modeling}

\author{
 \textbf{Ioannis Tsiamas\textsuperscript{1,2}}  \hspace{0.5cm}
 \textbf{David Dale\textsuperscript{1}}  \hspace{0.5cm}
 \textbf{Marta R. Costa-jussà\textsuperscript{1}}
\\
 \textsuperscript{1}FAIR at Meta, Paris \hspace{0.5cm}
 \textsuperscript{2}Universitat Politècnica de Catalunya, Barcelona
\\
 \small{
   \textbf{Correspondence:} \href{mailto:ytsiamas@meta.com}{ytsiamas@meta.com}, \href{mailto:daviddale@meta.com}{daviddale@meta.com}
 }
}

\begin{document}
\maketitle
\begin{abstract}
    Current translation systems, despite being highly multilingual, cover only 5$\%$ of the world's languages. Expanding language coverage to the long-tail of low-resource languages requires data-efficient methods that rely on cross-lingual and cross-modal knowledge transfer. To this end, we propose a character-based approach to improve adaptability to new languages and modalities. Our method leverages SONAR, a multilingual fixed-size embedding space with different modules for encoding and decoding. We use a teacher-student approach with parallel translation data to obtain a character-level encoder. Then, using ASR data, we train a lightweight adapter to connect a massively multilingual CTC ASR model (MMS), to the character-level encoder, potentially enabling speech translation from 1,000+ languages. Experimental results in text translation for 75 languages on FLORES+ demonstrate that our character-based approach can achieve better language transfer than traditional subword-based models, especially outperforming them in low-resource settings, and demonstrating better zero-shot generalizability to unseen languages. Our speech adaptation, maximizing knowledge transfer from the text modality, achieves state-of-the-art results in speech-to-text translation on the FLEURS benchmark on 33 languages, surpassing previous supervised and cascade models, albeit being a zero-shot model with minimal supervision from ASR data.
\end{abstract}

\section{Introduction}

    \begin{figure}[ht!]
        \centering
        \includegraphics[width=\columnwidth]{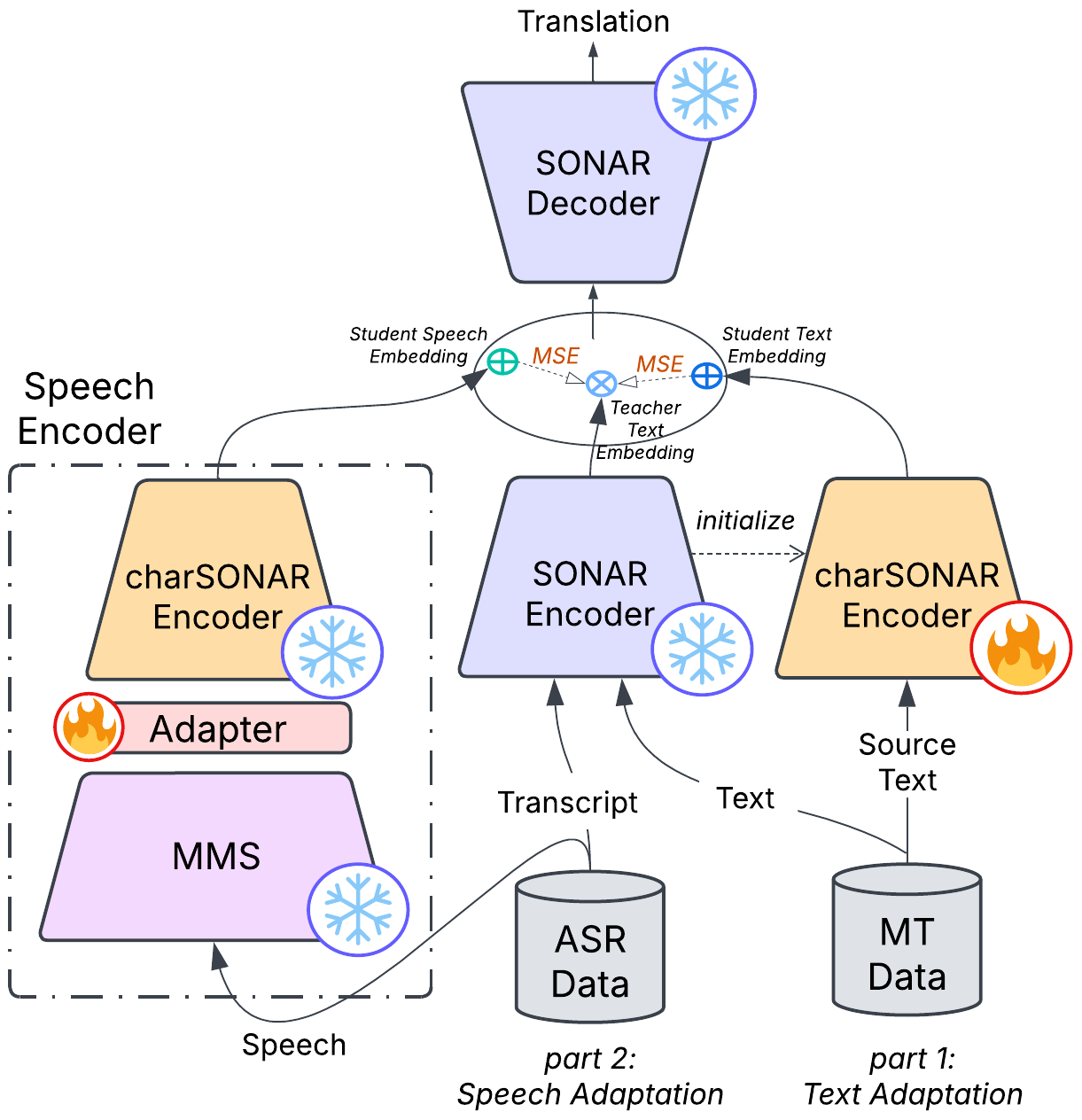}
         \caption{Approach for character-level and speech adaptation using the SONAR space.}
        \label{fig:intro}
    \end{figure}
    Translation has experienced a large growth in terms of language coverage in the last years, with models supporting 200-400 languages in text~\cite{nllb,madlad}, and 100 in speech~\cite{seamless}. Although impressive in terms of population coverage (90$\%$), in terms of actual language coverage we stand at only 5$\%$.\footnote{\href{https://www.ethnologue.com}{www.ethnologue.com}} Moving towards expanding to the long-tail of low-resource languages in the world posses some serious challenges due to the increasingly scarce data sources. For text translation we have to rely on a few thousand parallel sentences, while chances are there are no parallel data for speech translation (ST). To ease the issue of data scarcity in low-resource settings, multilinguality for text~\cite{google_mt,when_is_multilinguality_a_curse} and multimodality for speech~\cite{improving_st_by_understanding_and_learning} can usually be beneficial. But how can we increase cross-lingual and cross-modal knowledge transfer from high-resource languages and modalities? Recent research suggests that character-level models exhibit better cross-lingual transfer in text translation, especially in low-resource scenarios~\cite{are_character_level_translations_worth}. Furthermore, for speech translation, methods usually take advantage of a text-based encoder for semantic modeling~\cite{improving_st_by_understanding_and_learning,rethinking_and_imroving_multitask_learning_for_st}, but the subword-based tokenization is incompatible in terms of length and content with the acoustic representations, thus creating a \emph{modality gap} that hinders knowledge transfer. Previous research mitigates this by either using a phoneme-based text encoder~\cite{improving_st_by_understanding_and_learning,ctc_meets_ot} or converting the acoustic representations to subword-like units~\cite{zeroswot}. But a phonemized input degrades performance due to ambiguity\footnote{Homophones, loss of orthographic information, etc.} and furthermore phonemizers for 1000+ languages might be infeasible~\cite{mms_zero_shot}, while subword-based compression requires a substantial amount of data. To this end, we propose to shift towards character-based encoders, that could support data-efficient knowledge transfer both between languages and between text and speech. Our method is based on SONAR~\cite{sonar}, which is an encoder-decoder with a fixed-size semantic embedding space that supports 200 languages, and on MMS~\cite{mms}, which is a CTC-based ASR model that supports 1,000+ languages. Using a teacher-student approach, we obtain a character-based text encoder that embeds sentences in the SONAR space. Then, we propose an adapter that seamlessly connects the CTC output space of MMS to the character-level input space of our encoder, requiring minimal supervision from audio-transcription pairs (Fig.~\ref{fig:intro}). Our experimental results in 75 languages on FLORES+~\cite{nllb}, show that compared to traditional subword-based models, our multilingual character-level SONAR encoder exhibits better cross-lingual knowledge sharing between known languages and superior zero-shot generalizability to unseen languages. Furthermore, our speech adaptation of the character-based encoder, despite relying only on ASR data, can maximize knowledge transfer from text, and thus surpasses the previous best supervised system~\cite{seamless} and strong cascades with Whisper~\cite{whisper}, achieving new state-of-the-art in FLEURS~\cite{fleurs}.

\section{Relevant Research} \label{sec:relevant_research}


\subsection{Character-level MT}

    Early works in machine translation investigated character-level approaches due their advantages in understanding and generating rare and unseen words, handling noise, having smaller vocabularies, and being simpler due to the removal of subword tokenization~\cite{bpe}. Several methods using attention-based sequence-to-sequence models~\cite{seq2seq,additive_attention} showed that character-level MT can reach or surpass subword-based approaches~\cite{char_mt_ling,char_mt_costajussa,char_mt_lee,char_mt_cherry,char-level_decoder}. Later, \citet{byt5} showed that ByT5, encoder-decoder multilingual language model operating on bytes, is more robust to noise and performs better in spelling-sensitive tasks, than its subword-based counterpart mT5~\cite{t5}. ~\citet{are_character_level_translations_worth} finetuned the ByT5 and mT5 models for translation, and found that character-level modeling is particularity effective when parallel data are limited. \citet{why_dont_people_use_char_level} sought to answer why fully character-level MT has not been widely adopted, which was attributed to lower efficiency, and an inability to confirm previous findings that had been suggesting better domain and morphological generalization. In this work, we propose an encoder-only character-level approach~\cite{encoder_only_char} based on SONAR~\cite{sonar}, and study the benefits of cross-lingual transfer in a large group of 75 languages, both in low-resource and in zero-shot settings. Several works have proposed methods that alleviate the additional computational costs stemming from the longer sequences that character- or byte-level models need to process~\cite{canine,charformer,blt}. Since our approach adopts character-level modeling only on the encoder side, and due to the fixed-size embedding bottleneck of SONAR, the computational overhead is minimal, and thus we do not study any architecture-based changes in this work.

\subsection{Cross-modal Transfer in ST}

Speech translation models have traditionally relied on cross-modal knowledge transfer from the more resourceful task of text translation to improve performance. Several works achieved this by using a multitasking framework of MT and ST, where they share the text modules between the two tasks, and the semantic text encoder accepts either acoustic representations or text embeddings as inputs~\cite{bridging_the_modality_gap,cross_modal_progressive_training,a_general_multi_task,stemm}. Another line of work aims at bridging the modality gap by additionally minimizing the distance between the speech-text representations of the encoders~\cite{improving_st_by_understanding_and_learning,const,waco}. ZeroSwot~\cite{zeroswot} eliminated the dependency on parallel ST data, and relied only on minimizing the Wasserstein distance~\cite{ot} between the speech-text of representations of the encoders using ASR data. In our framework we also follow the paradigm of ZeroSwot, but due to the fixed-size encoder bottleneck of SONAR, our optimization is simpler, and minimizes the MSE distance. Another important consideration in maximizing knowledge transfer from text is unifying the tokenization of acoustic encoder's output and text encoder's input space, which are usually phoneme/characters for the the CTC~\cite{ctc} of the acoustic encoder, and subwords for the embedding layer of the text encoder. Previous works have either used a phoneme-based input for the text encoder~\cite{improving_st_by_understanding_and_learning,ctc_meets_ot}, or a subword-based output for the acoustic encoder's CTC output~\cite{bridging_the_modality_gap,towards_zero_shot}, or more recently a character-to-subword compression adapter~\cite{zeroswot}. But phoneme-based text input degrades performance due to ambiguity in meaning. Then, for subword-based output in CTC, it is questionable whether it can scale to massively multilingual vocabularies of hundreds of thousand of tokens~\cite{nllb}. Finally the subword compression adapter requires a substantial amount of ASR data to learn, which can be problematic for the long tail of low-resource languages. Contrary, our approach is based on first modifying the text encoder to work with character-level inputs without degrading MT performance, and then learning a data-efficient and lightweight adapter that connects to it the character-based output space of a CTC acoustic encoder.





\section{Methodology} \label{sec:methodology}

    We utilize the multilingual fixed-size embedding space of SONAR~\cite{sonar}, in order to add new languages and modalities (speech) to it. We first obtain a character-level text encoder using a teacher-student approach with parallel translation data~(\S\ref{sec:methodology_char_encoder}), and then adapt it to work with CTC acoustic representations as inputs using a teacher-student approach with paired audio-transcriptions~(\S\ref{sec:methodology_speech_encoder}).

    \subsection{SONAR} \label{sec:methodology_sonar}

        The SONAR encoder is a Transformer~\cite{attention_is_all_you_need} with $N_t$ layers of dimensionality $d_t$, and a subword-based vocabulary $\mathcal{V}_t$. The final encoder representation is mean-pooled to obtain a sentence embedding $\mathbf{e} \! \in \! \mathbb{R}^{d_t}$. The SONAR decoder which also has $N_t$ layers of dimensionality $d_t$, attends with cross-attention to $\mathbf{e}$, in order to predict the target sequence.

    \subsection{Character-level Text Encoder} \label{sec:methodology_char_encoder}

        \setlength{\parindent}{0pt}

        Our character-level encoder (charSONAR) is initialized from the SONAR encoder, and thus has $N_t$ layers of dimensionality $d_t$. As part of the character-based input vocabulary we only keep the tokens of $\mathcal{V}_t$ that are composed of single characters, thus having a vocabulary $\mathcal{V}_c \subset \mathcal{V}_t$.



        \textbf{Training Objectives.} For training, we follow a student-teacher approach with the SONAR encoder as a teacher, where we minimize the MSE loss between the charSONAR embedding $\mathbf{c}$ and a SONAR embedding $\mathbf{e} \in \mathbb{R}^d$, using monolingual or parallel translation data. We consider three different MSE objectives:

        \begin{itemize}
            \setlength{\itemsep}{0pt}
            \setlength{\parskip}{0pt} 
            \item \emph{Reconstruction}, where we learn from non-parallel data, and given a sentence $x$, we minimize $\mathcal{L}^{recon} = \text{MSE}(\mathbf{c}^x, \mathbf{e}^x)$.
            \item  \emph{Translation}, where we learn from parallel data, and given a sentence $x$ with translation $y$, we minimize $\mathcal{L}^{trans} = \text{MSE}(\mathbf{c}^x, \mathbf{e}^y)$.
            \item \emph{Interpolation}, where we also learn from parallel data, and given a sentence $x$ with translation $y$, we minimize the distance from the `average' teacher embedding for that pair (Eq.~\ref{eq:inteprol}).
        \end{itemize}
        \begin{align} \label{eq:inteprol}
            \mathcal{L}^{interpol} = \text{MSE}\left(\mathbf{c}^x, \frac{\mathbf{e}^x +\mathbf{e}^y}{2}\right)
        \end{align}
        \textbf{Augmentations.} We apply ASR-like augmentations to make the character-based encoder robust to the normalized and error-prone output of CTC ASR models and increase cross-modal transfer. Specifically with some probability $p^{norm}$, we normalize the source text input of the char-based encoder, removing casing and punctuation. Furthermore, with some small probability $p^{noise}$, we inject different noise perturbations to the text, such as character addition, deletion and replacement.


        \setlength{\parindent}{12pt}


    \subsection{Speech Encoder} \label{sec:methodology_speech_encoder}

        Our speech encoder is composed of an acoustic encoder, an adapter, and the charSONAR encoder.


        \subsubsection{Acoustic Encoder}

        The acoustic encoder consists of a series of strided convolutional layers, followed by a Transformer encoder with $N_s$ layers of dimensionality $d_s$. It is initialized from MMS~\cite{mms}, which was pretrained with the self-supervised objective of wav2vec 2.0~\cite{wav2vec2.0} and fine-tuned with CTC~\cite{ctc} on 1,000+ languages. Each language $i$ has its own CTC prediction head $\mathbf{W}^{(i)} \! \in \! \mathbb{R}^{d_s \times |\mathcal{B}_i|}$, where $\mathcal{B}_i$ is a language-specific character-based vocabulary (including the \texttt{<blank>} token), with $\mathcal{B}^{(i)} \subset \mathcal{V}_c$.
        
        The acoustic encoder is kept frozen during training, and with it we extract the final encoder representation $\mathbf{H} \in \mathbb{R}^{m \times d_s}$. Next, we apply CTC-based compression~\cite{ctc_compression} to remove redundancy and obtain a representation that is similar in length as the character-based tokenization of our charSONAR encoder. We label each point $j$ of $\mathbf{H}$ with its CTC prediction $\pi_j \! = \! \text{argmax}( \mathbf{W}^{(i)} \mathbf{h}_j)$, then average consecutive points corresponding to the same prediction, and drop points corresponding to \texttt{<blank>}. We thus obtain an acoustic representation $\mathbf{A} \! \in \! \mathbb{R}^{n \times d_s}$, where $n < m$.
        

        \subsubsection{Cross-modal Adapter}

        We use a cross-modal adapter to process the acoustic representation $\mathbf{A}$ into an embedding-like representation $\mathbf{E} \! \in \! \mathbb{R}^{n \times d_t}$, that aims to match as close as possible the character embedding expected at the input of the charSONAR encoder.

        To maximize pretrained knowledge and obtain an adapter that could work out-of-the box in extremely low-resource settings, we propose a minimal (pretrained) two-layer architecture that is fully initialized from MMS and charSONAR. Specially, we use the CTC classification layer $\mathbf{W}^{(i)}$ of MMS to project $\mathbf{A}$ to logits, and with a softmax we obtain a probability distribution over the MMS vocabulary $\mathcal{B}^{(i)}$. Then, since $\mathcal{B}^{(i)} \subset \mathcal{V}_{c}$, we can connect the two spaces by doing a soft prediction over the charSONAR vocabulary using its embedding layer.
        \begin{equation}
            \mathbf{E}^\text{pt} = \text{softmax} \left( \mathbf{A} \mathbf{W}^{(i)} \right) \mathbf{Emb}^{(i)}, \nonumber
        \end{equation}
        where $\mathbf{Emb}^{(i)} \! \in \! \mathbb{R}^{|\mathcal{B}^{(i)}| \times d_t}$ is the embedding layer of charSONAR, indexed by the entries of $\mathcal{B}^{(i)}$.
        
\begin{figure}[ht]
            \centering
            \includegraphics[width=\columnwidth]{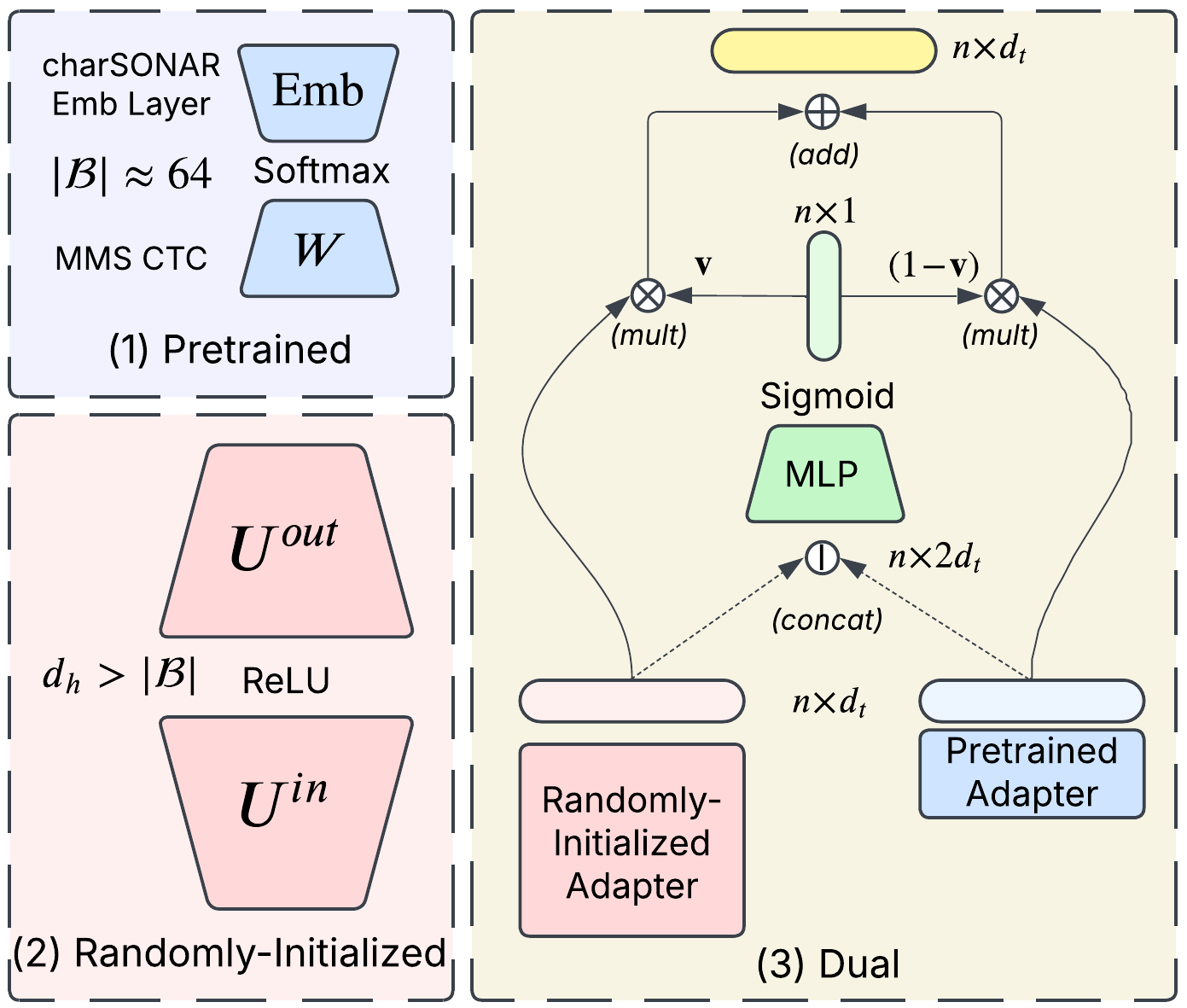}
             \caption{Cross-modal Adapters}
            \label{fig:adapter}
         \end{figure}

        Due to the nature of its initialization, the hidden dimension of the pretrained cross-modal adapter is fixed and bound to the size $|\mathcal{B}^{(i)}|$ of the MMS vocabulary, which is relatively small, usually having 64 tokens. In order to be able to control, and increase, the capacity of the adapter, we also propose a dual cross-modal adapter that combines the pretrained one with another variable-sized adapter that is randomly initialized (Fig.~\ref{fig:adapter}).
        \begin{equation}
            \mathbf{E}^\text{rnd} = \text{ReLU} \left( \mathbf{A} \mathbf{U}^{in} ) \right) \mathbf{U}^{out}, \nonumber
        \end{equation}
        where $\mathbf{E}^\text{rnd}$ is the output of the randomly-initialized adapter, $\mathbf{U}^{in} \in \mathbb{R}^{d_s \times d_h}$ and $\mathbf{U}^{out} \in \mathbb{R}^{d_h \times d_t}$ are learnable parameters, and $d_h$ is a hyperparameter that we can control. We concatenate the individual outputs of the pretrained and randomly-initialized adapters and pass them through an $\text{MLP}\!:\!2d_t\!\rightarrow\!1$, followed by a sigmoid function to obtain a vector of weights $\mathbf{v} \! \in \! (0, 1)^{n}$. The final representation $\mathbf{E}^\text{dual} \! \in \! \mathbb{R}^{n \times d_t}$ is a weighted sum of $\mathbf{E}^\text{pt}, \mathbf{E}^\text{rnd}$.
        \begin{align*}
            \mathbf{v} &= \sigma \left( \text{MLP} \left([\mathbf{E}^\text{pt}, \mathbf{E}^\text{rnd}] \right) \right) \\
            \mathbf{E}^\text{dual} &= \mathbf{v} \mathbf{E}^\text{pt} + (1-\mathbf{v}) \mathbf{E}^\text{rnd}
        \end{align*}

        To the output of the cross-modal adapter $\mathbf{E}$ we prepend the corresponding language token embedding, and append the embedding for end-of-sentence from the charSONAR embedding table. After adding positional encoding, $\mathbf{E}$ is passed through the transformer layers of the (frozen) charSONAR encoder to obtain a speech embedding $\mathbf{c}^z \! \in \! \mathbb{R}^{d_t}$. To train the adapter we use audio-transcription pairs, and minimize the MSE loss between $\mathbf{c}^z$ and the SONAR embedding for the transcription $\mathbf{e}^x$. For speech translation inference, we use the SONAR decoder to generate the translation from the speech embedding $\mathbf{c}^z$.


\section{Experimental Setup} \label{sec:exp_setup}

    \setlength{\parindent}{0pt}
    
    \subsection{Data} \label{sec:exp_setup_data}

        \textbf{Text.} We construct a diverse group of 63 languages in terms of family and script, and with varying degrees of resourcefulness, that are already present in SONAR. We also add a group of 12 new languages, not present in SONAR, for which we have evaluation data in FLORES+.\footnote{\href{https://huggingface.co/datasets/openlanguagedata/flores_plus}{huggingface.co/datasets/openlanguagedata/flores\_plus}} For the 63 known languages, to train charSONAR we used a combination of human-labeled data and mined parallel data~\cite{nllb}, which were filtered with BLASER 2.0~\cite{blaser2}, discarding pairs with score lower than 4. For the group of 12 new languages, we used various publicly available sources of parallel data. For validation and testing we used the \texttt{dev} and \texttt{devtest} splits of FLORES+.


        \textbf{Speech.} For our experiments in speech we used the 33 source languages. Our criteria for choosing these languages where: (a) part of the text training; (b) supported by MMS; (c) included in the Common Voice (CV)~\cite{cv} dataset (ASR training data); and (d) included in the FLEURS~\cite{fleurs} dataset (ST evaluation data). For training we used the \texttt{train} split of the version 17.0 of CommonVoice,\footnote{\href{https://huggingface.co/datasets/mozilla-foundation/common_voice_17_0}{datasets/mozilla-foundation/common\_voice\_17\_0}} and in some experiments the small \texttt{train} split of FLEURS which contains 2K examples for each language. Evaluation is done on the \texttt{dev} and \texttt{test} splits of FLEURS, which contain approximately 400 and 900 examples each.
        
        Details regarding the languages and the amount of training data for both text and speech are available in Table~\ref{tab:data} in the Appendix.

    \subsection{Model Architecture}  \label{sec:exp_setup_model}
    
        The SONAR encoder~\cite{sonar} has $N_t \! = \! 24$ layers, with dimensionality of $d_t \! = \! 1024$, and an embedding table of size 256K (750M parameters in total). Our charSONAR encoder follows the same architecture, apart from the character-based embedding table with a size of 8K tokens (500M parameters in total). MMS~\cite{mms} has $N_s \! = \! 48$ layers with dimensionality $d_s \! = \! 1280$ (1B parameters).\footnote{\href{https://huggingface.co/facebook/mms-1b-all}{huggingface.co/facebook/mms-1b-all}} It uses language-specific layers~\cite{adapters} and CTC classification layers. The vocabulary is different for each language, usually having around 64 tokens. Since the size of the vocabulary is also the hidden dimension of our pretrained adapter, this adapter has approximately 200K parameters. The randomly-initialized adapter uses a hidden dimension of $d_h\!=\!1024$ (2.2M parameters). For the dual adapter we use an MLP with an inner dimension of 64 (100K parameters) to predict the weight vector, thus having a total of 2.5M parameters. To generate translations, either from text or speech, we couple the encoder with the SONAR decoder which has 24 layers. For X$\rightarrow$Eng generation we use the normal SONAR decoder,\footnote{\href{https://dl.fbaipublicfiles.com/SONAR/sonar_text_decoder.pt}{dl.fbaipublicfiles.com/SONAR/sonar\_text\_decoder.pt}} while for all other generation tasks we use the finetuned decoder,\footnote{\href{https://dl.fbaipublicfiles.com/SONAR/finetuned_decoder.pt}{dl.fbaipublicfiles.com/SONAR/finetuned\_decoder.pt}} which according to \citet{sonar} and our observations here, performs better.

    \subsection{Training Details} \label{sec:exp_setup_training}
    
        \textbf{Text.} We use AdamW \cite{adamw} with a learning rate of 4e-4, inverse square root scheduler with warmup, a batch size of 12K examples, dropout of 0.1, and train for 128K steps. We up-sample languages with a temperature of 0.5~\cite{nllb}. We apply ASR-like text normalization by un-casing and removing punctuation to a source sentence with $p^{norm} \! = \! 0.25$, and inject character-based noise with $p^{noise}\!=\!0.125$. Specifically, each character in the source sentence can be deleted, replaced, or a new character is further added, each with a probability of 0.0025. These values were tuned in a small validation set of CV to approximate the character-error-rate of MMS~\cite{mms}. For replacement and addition we sample a new character from the character distribution of that language.

        \textbf{Speech.} Both MMS and charSONAR remain frozen during speech training, and only the adapter is finetuned. We minimize the MSE distance with the original SONAR as a teacher. The learning rate is set to 2e-4, the batch size 500 examples, and the adapter dropout to 0.1 (0.3 for the randomly-initialized adapter).
        
    \subsection{Evaluation}
    
        We apply checkpoint averaging according to the \texttt{dev} set performance, and generate with a beam search of 5. We evaluate primary on two tasks: translation and similarity search.
        Translation quality is measured with xCOMET-XL~\cite{xcomet}\footnote{\href{https://huggingface.co/Unbabel/XCOMET-XL}{huggingface.co/Unbabel/XCOMET-XL}}. When the target language is not supported, we use case-sensitive detokenized BLEU~\cite{sacrebleu} and chrF++~\cite{chrf2}. For similarity search we measure xSIM++~\cite{xsimplusplus} error rates, by augmenting the English parts of FLORES or FLEURS with 40K hard negatives.
    

    \setlength{\parindent}{12pt}

\section{Text Results} \label{sec:text_results}

        Here we present our results in text translation and similarity search with charSONAR, and investigate its capacity for cross-lingual knowledge transfer.

        \subsection{Initial Exploration} \label{subsec:text_initial_exploration}
        
        Before the main experiments we conduct an exploration regarding the training objectives and augmentations. We used a small subgroup of 15 languages, with 3 languages from the Uralic family, and 12 languages that use the Cyrillic script (Table~\ref{tab:data}). In the upper part of Table~\ref{tab:text_ablations} we show that the proposed interpolated MSE objective surpasses both the reconstruction and translation MSE objectives, and additionally their combination. This shows that SONAR embeds sentences in sub-optimal regions, while there are regions in between the languages that are better suited for both translation and similarity search. Further motivation is provided by our results of Table~\ref{tab:interpol_embedding_study}, where we show that for a pair of non-English languages $\text{Lang}_1$ and $\text{Lang}_2$, decoding from their average SONAR embedding $\text{Emb}_\text{AVG} \! = \! (\text{Emb}_1+\text{Emb}_2)/2$ into English, is better than decoding from each individual embedding, with low-resource languages benefiting the most.\footnote{High-resource are negatively impacted when paired with low-resource, but this reflects only a very small fraction of their data.} This finding indicates that our charSONAR encoder can benefit from learning to map sentences to the interpolated or `average' space existing between languages.
        
        In the lower part of Table~\ref{tab:text_ablations}, we find that pretraining charSONAR with the reconstruction MSE before the interpolated MSE is beneficial, since it decouples learning character-level modeling and optimizing the embedding space. Finally, we see that the normalization and noise augmentation do not have an impact in performance. This is expected due to the ground truth source text, but as we show later in the initial exploration for the speech experiments (\S\ref{subsec:speech_initial_exploration}), these augmentations are beneficial, as the input to charSONAR is error-prone.
        

        \begin{table}[ht]
            \centering
            \resizebox{\columnwidth}{!}{%
            \begin{tabular}{@{}cccccc@{}}
            \toprule
            \multicolumn{4}{c}{\textbf{Model}}                                      & \textbf{COMET} & \textbf{xSIM++} \\ \midrule
            \multicolumn{4}{c}{SONAR-200}                                           & 0.925          & 8.5             \\ \midrule
            \multicolumn{4}{c}{charSONAR-Ural/Cyrl}                                 &                &                 \\
            \textbf{Objective} & \textbf{Pretrain} & \textbf{Norm} & \textbf{Noise} &                &                 \\
            recon              & $\xmark$          & $\xmark$      & $\xmark$       & 0.929          & 7.4             \\
            trans              & $\xmark$          & $\xmark$      & $\xmark$       & 0.924          & 6.6             \\
            recon+trans        & $\xmark$          & $\xmark$      & $\xmark$       & 0.929          & 6.8             \\
            interpol           & $\xmark$          & $\xmark$      & $\xmark$       & 0.931          & 6.6             \\ \hdashline
            interpol           & $\cmark$          & $\xmark$      & $\xmark$       & \textbf{0.934} & \textbf{6.4}    \\
            interpol           & $\cmark$          & $\cmark$      & $\xmark$       & \textbf{0.934} & \textbf{6.4}    \\
            interpol           & $\cmark$          & $\cmark$      & $\cmark$       & \textbf{0.934} & 6.5             \\ \bottomrule
            \end{tabular}%
            }
             \caption{Ablations on training objectives and augmentations for the Ural/Cyrl language group (15 langs). Text translation COMET scores and cross-lingual xSIM++($\downarrow$) error rates on FLORES \texttt{dev} (X$\rightarrow$Eng).}
            \label{tab:text_ablations}
         \end{table}

        \begin{table}[ht]
            \centering
            \resizebox{\columnwidth}{!}{%
            \begin{tabular}{@{}ccccccc@{}}
            \toprule
            \multicolumn{2}{c}{\textbf{Pairs}} & \multicolumn{3}{c}{\textbf{COMET}} & \multicolumn{2}{c}{\textbf{Advantage}} \\ \midrule
            Lang$_1$ & Lang$_2$ & Emb$_1$ & Emb$_2$ & Emb$_{\text{AVG}}$ & Lang$_1$                       & Lang$_2$                       \\ \midrule
            Low      & Low      & 0.788   & 0.795   & 0.864              & \cellcolor[HTML]{DAE8FC}+0.076 & \cellcolor[HTML]{DAE8FC}+0.069 \\
            Low      & High     & 0.793   & 0.937   & 0.920              & \cellcolor[HTML]{34CDF9}+0.137 & \cellcolor[HTML]{FAE7E6}-0.017 \\
            High     & High     & 0.939   & 0.937   & 0.944              & \cellcolor[HTML]{ECF4FF}+0.005 & \cellcolor[HTML]{ECF4FF}+0.007 \\ \bottomrule
            \end{tabular}%
            }
             \caption{COMET scores of translating from average (interpolated) embeddings, compared to translating from individual embeddings, for different pairs based on resourcefulness. Results in X$\rightarrow$Eng FLORES \texttt{dev} averaged over 50 randomly-sampled pairs in each row.}
            \label{tab:interpol_embedding_study}
         \end{table}

        \begin{table*}[ht]  
            \centering
            \resizebox{0.8\textwidth}{!}{
            \begin{tabular}{@{}lcccccccccc@{}}
                    \toprule
                     &
                      \multicolumn{5}{c}{\textbf{COMET} ($\uparrow$)} 
                       &
                      \multicolumn{5}{c}{\textbf{xSIM++} ($\downarrow$)} \\ \midrule 
                    \textbf{Model} &
                      \begin{tabular}[c]{@{}c@{}}\textbf{Low}\\ (21)\end{tabular} &
                      \begin{tabular}[c]{@{}c@{}}\textbf{Med}\\ (21)\end{tabular} &
                      \begin{tabular}[c]{@{}c@{}}\textbf{High}\\ (21)\end{tabular} &
                      \begin{tabular}[c]{@{}c@{}}\textbf{All}\\ (63)\end{tabular} &
                      \begin{tabular}[c]{@{}c@{}}\textbf{New}\\ (12)\end{tabular} &
                      \begin{tabular}[c]{@{}c@{}}\textbf{Low}\\ (21)\end{tabular} &
                      \begin{tabular}[c]{@{}c@{}}\textbf{Med}\\ (21)\end{tabular} &
                      \begin{tabular}[c]{@{}c@{}}\textbf{High}\\ (21)\end{tabular} &
                      \begin{tabular}[c]{@{}c@{}}\textbf{All}\\ (63)\end{tabular} &
                      \begin{tabular}[c]{@{}c@{}}\textbf{New}\\ (12)\end{tabular} \\ \midrule
                    \multicolumn{10}{c}{\textit{Previous Works (trained on 200, not including the 12 new)}} \\
                    NLLB-200 &
                      0.877 &
                      \textbf{0.914} &
                      \textbf{0.949} &
                      0.913 &
                      0.454$^\dagger$ &
                      - &
                      - &
                      - &
                      - &
                      - \\
                    SONAR-200 &
                      0.851 &
                      0.894 &
                      0.944 &
                      0.897 &
                      0.450$^\dagger$ &
                      13.1 &
                      10.1 &
                      7.3 &
                      10.2 &
                      52.7$^\dagger$ \\
                      \midrule
                      \multicolumn{10}{c}{\textit{This work (trained on 63 known + 12 new languages)}} \\
                      SONAR-75 &
                      0.882 &
                      0.909 &
                      0.948 &
                      0.913 &
                      0.859 &
                      9.0 &
                      7.7 &
                      5.8 &
                      7.5 &
                      12.6 \\
                    charSONAR-75 &
                      \textbf{0.889} &
                      \textbf{0.914} &
                      \textbf{0.949} &
                      \textbf{0.917} &
                      \textbf{0.863} &
                      8.4 &
                      \textbf{7.2} &
                      \textbf{5.5} &
                      \textbf{7.0} &
                      \textbf{12.3} \\
                    \multicolumn{1}{r}{$\Delta$} &
                      \cellcolor[HTML]{34CDF9}0.007 &
                      \cellcolor[HTML]{DAE8FC}0.005 &
                      \cellcolor[HTML]{ECF4FF}0.001 &
                      \cellcolor[HTML]{DAE8FC}0.004 &
                      \cellcolor[HTML]{DAE8FC}0.004 &
                      \cellcolor[HTML]{34CDF9}0.6 &
                      \cellcolor[HTML]{DAE8FC}0.5 &
                      \cellcolor[HTML]{ECF4FF}0.3 &
                      \cellcolor[HTML]{DAE8FC}0.5 &
                      \cellcolor[HTML]{DAE8FC}0.3 \\
                      \midrule
                      \multicolumn{10}{c}{\textit{This work (trained only on the 63 known languages)}} \\
                    SONAR-63 &
                      0.882 &
                      0.909 &
                      0.947 &
                      0.913 &
                      0.517$^\dagger$ &
                      8.8 &
                      7.7 &
                      5.8 &
                      7.5 &
                      42.9$^\dagger$ \\
                    charSONAR-63 &
                      \textbf{0.889} &
                      \textbf{0.914} &
                      \textbf{0.949} &
                      \textbf{0.917} &
                      \underline{0.530}$^\dagger$ &
                      \textbf{8.3} &
                      \textbf{7.2} &
                      \textbf{5.5} &
                      \textbf{7.0} &
                      \underline{42.2}$^\dagger$ \\
                    \multicolumn{1}{r}{$\Delta$} &
                      \cellcolor[HTML]{34CDF9}0.007 &
                      \cellcolor[HTML]{DAE8FC}0.005 &
                      \cellcolor[HTML]{ECF4FF}0.002 &
                      \cellcolor[HTML]{DAE8FC}0.004 &
                      \cellcolor[HTML]{14bbeb}0.013 &
                      \cellcolor[HTML]{34CDF9}0.5 &
                      \cellcolor[HTML]{DAE8FC}0.5 &
                      \cellcolor[HTML]{ECF4FF}0.3 &
                      \cellcolor[HTML]{DAE8FC}0.5 &
                      \cellcolor[HTML]{14bbeb}0.7 \\
                      \bottomrule
                    \end{tabular}%
            }
            \caption{Translation COMET scores and cross-lingual similarity search xSIM++ error rates on FLORES \texttt{devtest} (X$\rightarrow$Eng), grouped by All(Low/Med/High) and New languages. $\dagger$ indicates zero-shot evaluation. \textbf{bold}: best overall; \underline{underlined}: best zero-shot. All models have the same number of parameters (1.3B). $\Delta$ refers to the difference between charSONAR-N and SONAR-N models.
            \label{tab:main_text_results}}
        \end{table*}

        \begin{table}[ht]
            \centering
            \resizebox{0.68\columnwidth}{!}{%
            \begin{tabular}{@{}lcc@{}}
            \toprule
            \textbf{Model} & \textbf{BLEU} & \textbf{chrF++} \\ \midrule 
            NLLB-200 & 17.4 & 45.3 \\ \midrule 
            SONAR-200      & 15.6          & 44.0                       \\
            SONAR-Eng      & \textbf{15.9} & \textbf{44.8}   \\
            charSONAR-Eng  & 15.8          & 44.7               \\ \bottomrule
            \end{tabular}%
            }
            \caption{Text translation (Eng$\rightarrow$200) BLEU and chrF++ scores on FLORES \texttt{devtest}.}
            \label{tab:eng_text_results}
        \end{table}
        
        \begin{table}[ht]
            \centering
            \resizebox{0.75\columnwidth}{!}{%
            \begin{tabular}{@{}lcc@{}}
            \toprule
            \textbf{Model} & \textbf{\# Tokens} & \textbf{\begin{tabular}[c]{@{}c@{}}Inference\\ Time (s)\end{tabular}} \\ \midrule
            SONAR          & 49\phantom{ ($\times$3.2)}                & 0.84\phantom{ ($\times$1.1)}                                                              \\
            charSONAR      & 158 ($\times$3.2) & 0.94 ($\times$1.1)                                               \\ \bottomrule
            \end{tabular}
             }
            \caption{Average number of tokens and average inference time in FLORES \texttt{dev}.}
            \label{tab:efficiency}
        \end{table}
        
        \subsection{Scaling to 75 Languages} \label{subsec:scaling_to_75}
        
        Next, we present our findings from scaling-up the language coverage of charSONAR to 75 languages. We use the interpolated MSE objective, with reconstruction MSE pretraining and ASR-like text augmentations. For new languages, which are not supported by SONAR, we use the translation MSE objective. We compare against SONAR-200~\cite{sonar}, and an NLLB-200~\cite{nllb} topline, which is not restricted by a bottleneck encoder representation. We also train a comparable subword-based model by further fine-tuning SONAR on the 75 languages with the same setup as we did for charSONAR. We report text translation and cross-lingual similarity search (X$\rightarrow$Eng) results, and group results by language resourcefulness according to the amount of our training data (Table~\ref{tab:data}). Our results of Table~\ref{tab:main_text_results} show the clear advantage of our character-based encoder, where charSONAR-75 outperforms the comparable SONAR-75, and additionally the NLLB topline in translation. The gains are more evident in the group of 21 low-resource languages, where cross-lingual transfer can be more impactful.

        \subsection{Zero-shot Generalization} \label{subsec:zero_shot_generalization}
        
        In our next experiment, we only train on the 63 known languages, and evaluate zero-shot on the 12 new ones. SONAR and NLLB encoders require a language tag to be prepended in the source sequence, which is problematic if we want to encode a sentence from a language not seen during training. To achieve better encoding for these unseen languages, we propose the use of family tokens according to the linguistic family subgroup of each language. Specifically, during training we replace the language token with the corresponding subgroup token with a 20$\%$ probability. On inference, we encode a new language, with the appropriate subgroup token. The subgroup tokens are trainable and are initialized from the average of the all the language tokens of each family.\footnote{Information about the linguistic families are available at Table \ref{tab:data} in the Appendix.} In the last part of Table~\ref{tab:main_text_results} we observe that charSONAR-63 can generalize better than a subword-based encoder to languages not seen during training, achieving an improvement of 0.013 points in COMET and 0.7 in xSIM++. We also notice a sharp increase for both our encoders, compared to original SONAR-200, showing the benefits of expanding language tokens to subgroup tokens.\footnote{For encoding a new language with SONAR-200 and NLLB-200, we do not use a language tag.}
        

        \subsection{Are the gains due to language transfer or more compute?} \label{subsec:gains_transfer_compute}
        
        An implicit side-effect of character-level modeling is that sequences are on average 3$\times$ longer, which means that the charSONAR encoder is using more FLOPs than the SONAR encoder. To further investigate the source of the advantage shown in Table~\ref{tab:main_text_results} we conduct an experiment where we train SONAR and charSONAR on only one language, specifically on English. The results of Table~\ref{tab:eng_text_results} show that in the single-language setting, there is no advantage for the character-based model, being slightly behind the subword-based one. This finding indicates that character-level modeling is beneficial due to better cross-lingual knowledge transfer, rather than due to increased compute.

        \subsection{Efficiency Analysis} \label{subsec:efficiency_analysis}
        
        To assess the degree of computational overhead due to the longer sequences, we measure the average inference time for the charSONAR and SONAR models in FLORES \texttt{dev} using a batch size of 1. The results of Table~\ref{tab:efficiency} show that although sequences are 3.2$\times$ longer for charSONAR, the inference time is only 1.1$\times$ longer. This is due to the encoder bottleneck, which decouples the decoder from the source sequence length. Results with batching are available in Table~\ref{tab:efficiency_batching} in the Appendix.

    \section{Speech Results} \label{sec:speech_results}

    To investigate the cross-modal benefits of character-level modeling, we present results in zero-shot speech translation and speech-text similarity search with our charSONAR-based speech encoder.
    
    

    \subsection{Initial Exploration} \label{subsec:speech_initial_exploration}

        In Table~\ref{tab:speech_ablations} we present zero-shot ST X$\rightarrow$Eng results in FLEURS \texttt{dev} for four languages of different families and of varying degree of resourcefulness, ranging from 3K examples (Estonian) to 330K (Spanish). We observe that the pretrained cross-modal adapter (\textsc{PreTr}), despite being significantly smaller, outperforms the large ($d_h \! = \! 1024$), yet randomly initialized, adapter (\textsc{Rnd}). Although for the high-resource Spanish, we notice that the difference is rather small, which indicates that with more data it can be beneficial to increase the capacity. Indeed, our proposed dual adapter (\textsc{Dual}), with large dimensionality in the random branch, surpasses them both. Finally, we notice further gains when we switch to a robust charSONAR version that was trained with ASR-like Norm/Noise augmentations~(\S\ref{sec:methodology_char_encoder}).

        \begin{table}[ht]
        \centering
        \resizebox{\columnwidth}{!}{%
        \begin{tabular}{@{\hskip 0.2cm}c@{\hskip 0.2cm}c@{\hskip 0.2cm}c@{\hskip 0.2cm}c@{\hskip 0.2cm}c@{\hskip 0.2cm}c@{\hskip 0.2cm}c@{\hskip 0.2cm}c@{\hskip 0.2cm}c@{\hskip 0.2cm}}
        \toprule
        \multicolumn{3}{c}{\textbf{Adapter}} & \textbf{Encoder}    & \multicolumn{5}{c}{\textbf{COMET}}             \\ \midrule
        \textbf{Type} & \textbf{Dim} & \textbf{Train} & \textbf{Norm/Noise} & \textbf{Est}   & \textbf{Rus}   & \textbf{Tur}   & \textbf{Spa}   & \textbf{Avg}   \\ \midrule
        \textsc{PreTr}     & $\sim$64    & $\xmark$    & $\xmark$ / $\xmark$ & 0.845 & 0.849 & 0.828 & 0.837          & 0.840 \\
        \textsc{PreTr}     & $\sim$64    & $\cmark$    & $\xmark$ / $\xmark$ & 0.901 & 0.910 & 0.877 & 0.890          & 0.894 \\ \midrule
        \textsc{Rnd}      & 256         & $\cmark$    & $\xmark$ / $\xmark$ & 0.837 & 0.872 & 0.831 & 0.878          & 0.854 \\
        \textsc{Rnd}      & 1024        & $\cmark$    & $\xmark$ / $\xmark$ & 0.882 & 0.889 & 0.869 & 0.889          & 0.882 \\ \midrule
        \textsc{Dual}     & 256         & $\cmark$    & $\xmark$ / $\xmark$ & 0.914 & 0.912 & 0.889 & 0.888          & 0.901 \\
        \textsc{Dual}     & 1024        & $\cmark$    & $\xmark$ / $\xmark$ & 0.911 & 0.909 & 0.894 & \textbf{0.905} & 0.905 \\
        \textsc{Dual}     & 1024        & $\cmark$    & $\cmark$ / $\xmark$ & 0.914 & 0.910 & 0.891 & 0.897          & 0.903 \\
        \textsc{Dual}          & 1024         & $\cmark$       & $\cmark$ / $\cmark$ & \textbf{0.915} & \textbf{0.923} & \textbf{0.906} & \textbf{0.905} & \textbf{0.912} \\ \bottomrule
        \end{tabular}%
        }
         \caption{Ablations in speech adaptation. Speech Translation (X$\rightarrow$Eng) results on FLEURS \texttt{dev}.}
        \label{tab:speech_ablations}
         \end{table}

        \subsection{Zero-shot Speech Translation} \label{subsec:zero_shot_speech_translation}
        
        Next, we train adapters for 33 languages using the charSONAR-75 encoder, and compare against strong supervised E2E models, Whisper~\cite{whisper} and SeamlessM4T~\cite{seamless}, cascades with MMS/Whisper and NLLB/SONAR, and our own cascades with SONAR-75/charSONAR-75. We report results by grouping languages according to number of examples in CommonVoice (CV) (Table~\ref{tab:main_speech_results}). The first version of our system~(11) using the pretrained adapter, can work out-of-the-box and without any training, even outperforming Whisper by a large margin, and particularly for low-resource languages. This indicates that the input space of our character-based encoder is fully compatible with the output space of MMS given the initialization of our adapter. Following, by training this adapter with ASR data (12), we surpass the previous state-of-the-art SeamlessM4T-Large-v2. The benefits of cross-modal transfer from charSONAR are evident for some extremely low-resource languages such as Asturian, where with only 400 examples it surpasses SeamlessM4T by 0.1 COMET (Table~\ref{tab:speech_analytic} in Appendix). Furthermore, we observe additional gains when using the proposed dual adapter~(13) for medium/high-resource languages, where there are enough data to learn the large, but randomly initialized, branch. Finally, we show that by adding only 2K additional examples from FLEURS \texttt{train}, we can achieve further important gains across all categories~(14-15).
        
        Apart from the strong cross-modal transfer showcased by our speech adaptation of charSONAR, significant gains are also observed for the cascade systems that employ it. Specifically a cascade of MMS and charSONAR~(9) outperforms all other cascades~(3-8) and is on par with SeamlessM4T-Large-v2 in low/medium-resource settings.

        \begin{table*}
            \centering
            \resizebox{0.94\textwidth}{!}{%
            \begin{tabular}{@{}clcccccccc@{}}
            \toprule
            \textbf{id} &
              \textbf{Model} &
              \textbf{\begin{tabular}[c]{@{}c@{}}Text\\ Encoder\\ Tokenization\end{tabular}} &
              \textbf{\begin{tabular}[c]{@{}c@{}}Total\\ Params\end{tabular}} &
              \textbf{\begin{tabular}[c]{@{}c@{}}Adapter\\ Train\\ Params\end{tabular}} &
              \textbf{\begin{tabular}[c]{@{}c@{}}Adapter\\ Train\\ Data\end{tabular}} &
              \textbf{\begin{tabular}[c]{@{}c@{}}Low\\ (11)\\ \end{tabular}} &
              \textbf{\begin{tabular}[c]{@{}c@{}}Med\\ (11)\\ \end{tabular}} &
              \textbf{\begin{tabular}[c]{@{}c@{}}High\\ (11)\\ \end{tabular}} &
              \textbf{\begin{tabular}[c]{@{}c@{}}All\\ (33)\\ \end{tabular}}
              \\ \midrule
            \multicolumn{2}{l}{\textbf{Supervised E2E ST (previous)}} \\
                1 &
              \textsc{Whisper-Large}-v3 &
              / &
              1.5B &
              / &
              / &
              0.598 &
              0.754 &
              0.790 &
              0.714 \\
            2 &
              \textsc{SeamlessM4T-Large}-v2 &
              / &
              2.3B &
              / &
              / &
              \underline{0.829} &
              \underline{0.889} &
              \underline{0.901} &
              \underline{0.873} \\ \midrule
            \multicolumn{2}{l}{\textbf{Cascade ST (previous)}} \\
            3 &
              MMS + NLLB-200 &
              subwords &
              2.3B &
              / &
              / &
              0.786 &
              0.834 &
              0.822 &
              0.814 \\
            4 &
              \textsc{Whisper} + NLLB-200 &
              subwords &
              2.8B &
              / &
              / &
              0.717 &
              0.870 &
              0.863 &
              0.817 \\
            5 &
              MMS + SONAR-200 &
              subwords &
              2.3B &
              / &
              / &
              0.757 &
              0.839 &
              0.824 &
              0.807 \\
            6 &
              \textsc{Whisper} + SONAR-200 &
              subwords &
              2.8B &
              / &
              / &
              0.684 &
              0.869 &
              0.861 &
              0.804 \\ \midrule
            \multicolumn{2}{l}{\textbf{Cascade ST (ours)}} \\
            7 &
              MMS + SONAR-75 &
              subwords &
              2.3B &
              / &
              / &
              0.811 &
              0.870 &
              0.854 &
              0.845 \\
            8 &
              \textsc{Whisper} + SONAR-75 &
              subwords &
              2.8B &
              / &
              / &
              0.721 &
              0.871 &
              0.865 &
              0.819 \\
            9 &
              MMS + charSONAR-75 &
              chars &
              2.3B &
              / &
              / &
              \cellcolor[HTML]{DAE8FC}0.833 &
              \cellcolor[HTML]{DAE8FC}0.889 &
              0.875 &
              0.866 \\
            10 &
              \textsc{Whisper} + charSONAR-75 &
              chars &
              2.8B &
              / &
              / &
              0.755 &
              \cellcolor[HTML]{DAE8FC}0.893 &
              0.882 &
              0.843 \\ \midrule
            \multicolumn{2}{l}{\textbf{Zero-shot E2E ST (ours)}} \\
            11 &
              Speech-charSONAR-75 - \textsc{PreTr} &
              chars &
              2.3B &
              0 &
              / &
              0.772 &
              0.833 &
              0.831 &
              0.812 \\
            12 &
              Speech-charSONAR-75 - \textsc{PreTr} &
              chars &
              2.3B &
              0.2M &
              CV &
              \cellcolor[HTML]{DAE8FC}0.837 &
              \cellcolor[HTML]{DAE8FC}0.893 &
              0.894 &
              \cellcolor[HTML]{DAE8FC}0.875 \\
            13 &
              Speech-charSONAR-75 - \textsc{Dual} &
              chars &
              2.3B &
              2.5M &
              CV &
              0.615 &
              \cellcolor[HTML]{DAE8FC}0.899 &
              \cellcolor[HTML]{DAE8FC}0.902 &
              0.805 \\
            14 &
              Speech-charSONAR-75 - \textsc{PreTr} &
              chars &
              2.3B &
              0.2M &
              CV+FLEURS &
              \cellcolor[HTML]{DAE8FC}0.852 &
              \cellcolor[HTML]{DAE8FC}0.900 &
              \cellcolor[HTML]{DAE8FC}0.901 &
              \cellcolor[HTML]{DAE8FC}0.884 \\
            15 &
              Speech-charSONAR-75 - \textsc{Dual} &
              chars &
              2.3B &
              2.5M &
              CV+FLEURS &
              \cellcolor[HTML]{DAE8FC}\textbf{0.853} &
              \cellcolor[HTML]{DAE8FC}\textbf{0.906} &
              \cellcolor[HTML]{DAE8FC}\textbf{0.910} &
              \cellcolor[HTML]{DAE8FC}\textbf{0.889} \\ \bottomrule
            \end{tabular}%
            }
             \caption{Speech Translation (33 $\rightarrow$ Eng) COMET scores on FLEURS \texttt{test}. Low/Med/High each contain 11 languages, according to amount of Common Voice data. \underline{Underlined} are the previous best scores. \highlight{Highlighted} are our scores with char-based models that are at least on par with the previous best. In \textbf{bold} are best overall.}
            \label{tab:main_speech_results}
        \end{table*}

    \subsection{Similarity Search} \label{subsec:similarity_search}
    
    In Table~\ref{tab:xsim_scores} we present results on cross-lingual and cross-modal similarity search on FLEURS \texttt{test}. We compare our character-based speech encoder against several cascades and the original SONAR speech encoders~\cite{sonar} that are based on w2v-BERT~\cite{w2v_bert}, and thus do not transfer knowledge from an acoustic model (MMS) nor the text modality (charSONAR). We observe that our minimal adapters trained on CV outperform the previous SONAR speech encoders and all cascades apart from the charSONAR-based one. Still, by using more data (FLEURS \texttt{train}) and/or more parameters (dual adapter), the proposed encoder surpasses the charSONAR cascades.

    \begin{table}
        \centering
        \resizebox{0.92\columnwidth}{!}{%
        \begin{tabular}{@{}lcc@{}}
        \toprule
        \textbf{Model} & \textbf{avg26} & \textbf{avg33} \\ \midrule
        SONAR Speech \cite{sonar} & 14.3 & / \\ \midrule
        MMS + SONAR-200 & 15.7 & 17.2 \\
        MMS + SONAR-75 & 12.7 & 13.8 \\
        MMS + charSONAR-75 & 10.7 & 11.5 \\ \midrule
        Speech-charSONAR-75 - \textsc{PreTr} & 11.6 & 12.9 \\
        $\hookrightarrow$ w/ FLEURS \texttt{train} & 10.0 & 10.7 \\
        $\phantom{aa} \hookrightarrow$ w/ \textsc{Dual} & \textbf{9.4} & \textbf{10.2} \\ \bottomrule
        \end{tabular}%
        }
         \caption{Cross-modal and cross-lingual retrieval. xSIM++ error rates ($\downarrow$) on FLEURS \texttt{test} (X$\rightarrow$Eng). avg26 is the languages supported by SONAR Speech \cite{sonar} and our models.}
        \label{tab:xsim_scores}
     \end{table}

    \subsection{Adapters for Subword-based Encoders} \label{subsec:adapters_for_subword}
    
    To understand how essential is the character-based encoder in our proposed speech adaptation, we experiment with replacing it with a subword-based one. To achieve this we mean-pool the indices of the compressed acoustic representation of MMS that belong to the same subword, as predicted by the CTC. The pretrained adapter version is not possible in this setting and thus we experiment only with the randomly-initialized adapter. The results of Table~\ref{tab:sw_adapter} indicate that we can learn an adapter to connect MMS and (subword-based) SONAR, although the quality is limited, and only works in high resource settings (Turkish, Spanish), while still being several points behind the character-based model. This highlights both the data-efficiency and cross-modal adaptability of our proposed method.

        \begin{table}
        \centering
        \resizebox{\columnwidth}{!}{%
        \begin{tabular}{@{}lcccc@{}}
        \toprule
        \textbf{Model} &
          \begin{tabular}[c]{@{}c@{}}\textbf{Oci}\\ \# 0.3k\end{tabular} &
          \begin{tabular}[c]{@{}c@{}}\textbf{Est}\\ \# 3k\end{tabular} &
          \begin{tabular}[c]{@{}c@{}}\textbf{Tur}\\ \# 30k\end{tabular} &
          \begin{tabular}[c]{@{}c@{}}\textbf{Spa}\\ \# 330k\end{tabular} \\ \midrule
        Speech-SONAR-75 - \textsc{Rnd}     & 0.199          & 0.223          & 0.795          & 0.841          \\
        Speech-charSONAR-75 - \textsc{Rnd} & \underline{0.202}    & \underline{0.877}    & \underline{0.868}    & \underline{0.912}    \\ \midrule
        Speech-charSONAR-75 - \textsc{PreTr}     & \textbf{0.795} & 0.910          & 0.885          & 0.917          \\
        Speech-charSONAR-75 - \textsc{Dual}    & 0.707          & \textbf{0.912} & \textbf{0.903} & \textbf{0.920} \\ \bottomrule
        \end{tabular}%
        }
         \caption{Speech Translation COMET (X$\rightarrow$Eng) on FLEURS \texttt{test} for subword-based vs char-based encoders with adapters. \underline{Underlined}: best among \textsc{Rnd} adapters; \textbf{bold}: best overall; $\#$: ASR examples.}
        \label{tab:sw_adapter}
         \end{table}

\section{Conclusions} \label{sec:conclusions}

    We presented a methodology based on character-level modeling that increases cross-lingual transfer and cross-modal transfer in text and speech tasks. For text, our character-based encoder surpasses comparable subword-based encoders, especially in low-resource settings, while exhibiting better zero-shot generalization to unseen languages. For speech, our proposed minimal adapter seamlessly connects an ASR CTC encoder to our character-based encoder, surpassing previous state-of-the-art models. Furthermore it requires minimal supervision from ASR data, and can even work out-of-the-box without any training, surpassing models like Whisper. Future research will focus on target-side cross-lingual and cross-modal transfer, and expanding to more languages.
    

\newpage

\section*{Limitations}

In this work we focused on source-side cross-lingual and cross-modal transfer, leaving target-side transfer for future research. We hypothesize that character-level modeling can be beneficial for target-side, although decoding on the character-level can be problematic and relatively more inefficient than encoding on the character-level. We still believe this is an interesting direction for future work.

Furthermore, we decided to focus on adapting a specific model, SONAR, to work with character-level input. Although the encoder bottleneck reduces the computational overhead in generation, and allowed us to simplify the teacher-student training by using an MSE objective, it also reduces the capacity of the model. We hypothesize that similar gains can be achieved by adapting a traditional encoder-decoder, like NLLB~\cite{nllb}, to work with characters, either by back-propagating the translation signal through the (frozen) decoder or using similar objectives to ZeroSwot~\cite{zeroswot}. Also, as discussed in our Relevant Research (\S\ref{sec:relevant_research}), we did not experiment with any specific architectural changes in the encoder that are better suited for character-level modeling~\cite{canine,charformer,blt}, as we aimed to study character-based vs subword-based modeling within the same architecture. We believe that by using such techniques further gains in performance and efficiency can be achieved. 

Additionally, our proposed methodology for speech adaptation is limited by the language-specific CTC layers of MMS. This forced us to train language-specific cross-modal adapters, which does not allow the speech encoder to generalize to more languages, other than the ones for which we have ASR data. To go around this issue, we carried some experiments with the zero-shot version of MMS~\cite{mms_zero_shot} that uses a unified model for all languages, but due to decreased ASR quality compared to MMS-1B~\cite{mms}, translation quality was also lagging behind. Still, in the future, and given a supervised MMS-like acoustic model with a unified architecture, our proposed cross-modal adapter could enable generalized speech understand and translation with it.


\section*{Acknowledgments}

We thank Pere Lluís Huguet Cabot for his helpful comments and suggestions.

\bibliography{custom}
\bibliographystyle{acl_natbib}

\appendix

\section{Data} \label{sec:appendix_data}

    In Table~\ref{tab:data} we provide details about the languages and the amounts of data used in our experiments. The numbers for the MT data indicate the amount after filtering with BLASER 2.0~\cite{blaser2}. The resourcefulness label (low, medium, or high) of each language is separate for each modality, and indicates in which of the three percentile of the data distribution it belongs. 
    

    \begin{table*}
        \centering
        \resizebox{\textwidth}{!}{%
\begin{tabular}{@{}ccccccccccc@{}}
\toprule
\multirow{2}{*}{\textbf{Language}} &
  \multirow{2}{*}{\textbf{Code}} &
  \multirow{2}{*}{\textbf{FLORES+}} &
  \multirow{2}{*}{\textbf{MMS}} &
  \multirow{2}{*}{\textbf{Family}} &
  \multirow{2}{*}{\textbf{Subgrouping}} &
  \multirow{2}{*}{\textbf{Script}} &
  \multicolumn{2}{c}{\textbf{MT}} &
  \multicolumn{2}{c}{\textbf{ASR}} \\ \cmidrule(l){8-11} 
 &
   &
   &
   &
   &
   &
   &
  \multicolumn{2}{c}{\textbf{\# (M) / Resource}} &
  \multicolumn{2}{c}{\textbf{\# (K) / Resource}} \\ \midrule
Aragonese         & arg\_Latn           & $\cmark$  & $\xmark$ & Indo-European   & Italic          & Latin      & 0.1   & new  & 0.0    & -    \\
Asturian          & ast\_Latn           & $\xmark$ & $\cmark$  & Indo-European   & Italic          & Latin      & 0.2   & low  & 0.4    & -    \\
Awadhi            & awa\_Deva           & $\xmark$ & $\cmark$  & Indo-European   & Indo-Aryan      & Devanagari & 0.4   & low  & 0.0    & -    \\
South Azerbaijani & azb\_Arab           & $\xmark$ & $\cmark$  & Turkic          & Common Turkic   & Arabic     & 0.3   & low  & 0.0    & -    \\
North Azerbaijani & azj\_Latn           & $\xmark$ & $\cmark$  & Turkic          & Common Turkic   & Latin      & 9.4   & med  & 0.1    & low  \\
Bashkir           & bak\_Cyrl           & $\xmark$ & $\cmark$  & Turkic          & Common Turkic   & Cyrillic   & 1.7   & med  & 119.2  & -    \\
Belarusian        & bel\_Cyrl           & $\xmark$ & $\cmark$  & Indo-European   & Balto-Slavic    & Cyrillic   & 11.5  & med  & 347.6  & high \\
Bhojpuri          & bho\_Deva           & $\xmark$ & $\xmark$ & Indo-European   & Indo-Aryan      & Devanagari & 0.6   & low  & 0.0    & -    \\
Bosnian           & bos\_Latn           & $\xmark$ & $\cmark$  & Indo-European   & Balto-Slavic    & Latin      & 21.8  & high & 0.0    & -    \\
Boro              & brx\_Deva           & $\cmark$  & $\xmark$ & Sino-Tibetan    & Tibeto-Burman   & Devanagari & 0.1   & new  & 0.0    & -    \\
Bulgarian         & bul\_Cyrl           & $\xmark$ & $\cmark$  & Indo-European   & Balto-Slavic    & Cyrillic   & 39.3  & high & 4.8    & med  \\
Catalan           & cat\_Latn           & $\xmark$ & $\cmark$  & Indo-European   & Italic          & Latin      & 10.1  & med  & 1146.2 & high \\
Valencian         & cat\_Latn\_vale1252 & $\cmark$  & $\xmark$ & Indo-European   & Italic          & Latin      & 0.0   & new  & 0.0    & -    \\
Czech             & ces\_Latn           & $\xmark$ & $\cmark$  & Indo-European   & Balto-Slavic    & Latin      & 52.3  & high & 20.1   & med  \\
Chuvash           & chv\_Cyrl           & $\cmark$  & $\cmark$  & Turkic          & Oghuric         & Cyrillic   & 1.2   & new  & 1.4    & -    \\
Central Kurdish   & ckb\_Arab           & $\xmark$ & $\cmark$  & Indo-European   & Iranian         & Arabic     & 1.7   & med  & 7.7    & -    \\
Crimean Tatar     & crh\_Latn           & $\xmark$ & $\cmark$  & Turkic          & Common Turkic   & Latin      & 0.2   & low  & 0.0    & -    \\
Dogri             & dgo\_Deva           & $\cmark$  & $\cmark$  & Indo-European   & Indo-Aryan      & Devanagari & 0.1   & new  & 0.0    & -    \\
Greek             & ell\_Grek           & $\xmark$ & $\cmark$  & Indo-European   & Graeco-Phrygian & Greek      & 52.6  & high & 1.9    & low  \\
Estonian          & est\_Latn           & $\xmark$ & $\cmark$  & Uralic          & Finnic          & Latin      & 16.9  & high & 3.2    & med  \\
Finnish           & fin\_Latn           & $\xmark$ & $\cmark$  & Uralic          & Finnic          & Latin      & 32.6  & high & 2.1    & low  \\
French            & fra\_Latn           & $\xmark$ & $\cmark$  & Indo-European   & Italic          & Latin      & 144.9 & high & 558.1  & high \\
Friulian          & fur\_Latn           & $\xmark$ & $\xmark$ & Indo-European   & Italic          & Latin      & 0.2   & low  & 0.0    & -    \\
Galician          & glg\_Latn           & $\xmark$ & $\cmark$  & Indo-European   & Italic          & Latin      & 6.5   & med  & 25.2   & high \\
Konkani           & gom\_Deva           & $\cmark$  & $\xmark$ & Indo-European   & Indo-Aryan      & Devanagari & 0.1   & new  & 0.0    & -    \\
Hindi             & hin\_Deva           & $\xmark$ & $\cmark$  & Indo-European   & Indo-Aryan      & Devanagari & 35.6  & high & 4.7    & med  \\
Chhattisgarhi     & hne\_Deva           & $\xmark$ & $\cmark$  & Indo-European   & Indo-Aryan      & Devanagari & 0.3   & low  & 0.0    & -    \\
Croatian          & hrv\_Latn           & $\xmark$ & $\cmark$  & Indo-European   & Balto-Slavic    & Latin      & 17.1  & high & 0.0    & -    \\
Hungarian         & hun\_Latn           & $\xmark$ & $\cmark$  & Uralic          & –               & Latin      & 32.6  & high & 37.1   & high \\
Italian           & ita\_Latn           & $\xmark$ & $\cmark$  & Indo-European   & Italic          & Latin      & 95.7  & high & 169.8  & high \\
Karakalpak        & kaa\_Latn           & $\cmark$  & $\cmark$  & Turkic          & Kipchak         & Latin      & 0.3   & new  & 0.0    & -    \\
Kashmiri          & kas\_Deva           & $\xmark$ & $\xmark$ & Indo-European   & Indo-Aryan      & Devanagari & 0.1   & low  & 0.0    & -    \\
Kazakh            & kaz\_Cyrl           & $\xmark$ & $\cmark$  & Turkic          & Common Turkic   & Cyrillic   & 5.6   & med  & 0.5    & low  \\
Halh Mongolian    & khk\_Cyrl           & $\xmark$ & $\cmark$  & Mongolic-Khitan & Mongolic        & Cyrillic   & 0.5   & low  & 2.2    & low  \\
Kyrgyz            & kir\_Cyrl           & $\xmark$ & $\cmark$  & Turkic          & Common Turkic   & Cyrillic   & 2.7   & med  & 1.8    & -    \\
Northern Kurdish  & kmr\_Latn           & $\xmark$ & $\cmark$  & Indo-European   & Iranian         & Latin      & 0.7   & med  & 5.1    & -    \\
Ligurian          & lij\_Latn           & $\xmark$ & $\xmark$ & Indo-European   & Italic          & Latin      & 0.2   & low  & 1.6    & -    \\
Lithuanian        & lit\_Latn           & $\xmark$ & $\cmark$  & Indo-European   & Balto-Slavic    & Latin      & 14.0  & high & 7.3    & med  \\
Lombard           & lmo\_Latn           & $\xmark$ & $\xmark$ & Indo-European   & Italic          & Latin      & 0.3   & low  & 0.0    & -    \\
Latgalian         & ltg\_Latn           & $\xmark$ & $\xmark$ & Indo-European   & Balto-Slavic    & Latin      & 0.3   & low  & 3.7    & -    \\
Standard Latvian  & lvs\_Latn           & $\xmark$ & $\cmark$  & Indo-European   & Balto-Slavic    & Latin      & 2.8   & med  & 11.4   & med  \\
Magahi            & mag\_Deva           & $\xmark$ & $\cmark$  & Indo-European   & Indo-Aryan      & Devanagari & 0.3   & low  & 0.0    & -    \\
Maithili          & mai\_Deva           & $\xmark$ & $\cmark$  & Indo-European   & Indo-Aryan      & Devanagari & 0.4   & low  & 0.0    & -    \\
Marathi           & mar\_Deva           & $\xmark$ & $\cmark$  & Indo-European   & Indo-Aryan      & Devanagari & 11.8  & med  & 2.2    & low  \\
Meadow Mari       & mhr\_Cyrl           & $\cmark$  & $\cmark$  & Uralic          & Finno-Ugric     & Cyrillic   & 0.4   & new  & 185.9  & -    \\
Macedonian        & mkd\_Cyrl           & $\xmark$ & $\cmark$  & Indo-European   & Balto-Slavic    & Cyrillic   & 6.8   & med  & 1.7    & low  \\
Erzya             & myv\_Cyrl           & $\cmark$  & $\cmark$  & Uralic          & Mordvinic       & Cyrillic   & 0.1   & new  & 1.2    & -    \\
Nepali            & npi\_Deva           & $\xmark$ & $\cmark$  & Indo-European   & Indo-Aryan      & Devanagari & 4.5   & med  & 0.3    & low  \\
Occitan           & oci\_Latn           & $\xmark$ & $\cmark$  & Indo-European   & Italic          & Latin      & 0.2   & low  & 0.3    & low  \\
Aranese           & oci\_Latn\_aran1260 & $\cmark$  & $\xmark$ & Indo-European   & Italic          & Latin      & 0.0   & new  & 0.0    & -    \\
Southern Pashto   & pbt\_Arab           & $\xmark$ & $\xmark$ & Indo-European   & Iranian         & Arabic     & 0.9   & med  & 0.0    & -    \\
Western Persian   & pes\_Arab           & $\xmark$ & $\cmark$  & Indo-European   & Iranian         & Arabic     & 15.0  & high & 28.9   & high \\
Polish            & pol\_Latn           & $\xmark$ & $\cmark$  & Indo-European   & Balto-Slavic    & Latin      & 60.4  & high & 20.7   & med  \\
Portuguese        & por\_Latn           & $\xmark$ & $\cmark$  & Indo-European   & Italic          & Latin      & 116.8 & high & 22.0   & med  \\
Dari              & prs\_Arab           & $\xmark$ & $\xmark$ & Indo-European   & Iranian         & Arabic     & 0.9   & med  & 0.0    & -    \\
Romanian          & ron\_Latn           & $\xmark$ & $\cmark$  & Indo-European   & Italic          & Latin      & 59.9  & high & 5.1    & med  \\
Russian           & rus\_Cyrl           & $\xmark$ & $\cmark$  & Indo-European   & Balto-Slavic    & Cyrillic   & 89.0  & high & 26.4   & high \\
Sanskrit          & san\_Deva           & $\xmark$ & $\xmark$ & Indo-European   & Indo-Aryan      & Devanagari & 0.3   & low  & 0.0    & -    \\
Sicilian          & scn\_Latn           & $\xmark$ & $\xmark$ & Indo-European   & Italic          & Latin      & 0.2   & low  & 0.0    & -    \\
Slovak            & slk\_Latn           & $\xmark$ & $\cmark$  & Indo-European   & Balto-Slavic    & Latin      & 29.7  & high & 3.3    & med  \\
Slovenian         & slv\_Latn           & $\xmark$ & $\cmark$  & Indo-European   & Balto-Slavic    & Latin      & 20.8  & high & 1.4    & low  \\
Sindhi            & snd\_Deva           & $\cmark$  & $\cmark$  & Indo-European   & Indo-Aryan      & Devanagari & 0.0   & new  & 0.0    & -    \\
Spanish           & spa\_Latn           & $\xmark$ & $\cmark$  & Indo-European   & Italic          & Latin      & 202.0 & high & 336.8  & high \\
Sardinian         & srd\_Latn           & $\xmark$ & $\xmark$ & Indo-European   & Italic          & Latin      & 0.2   & low  & 0.5    & -    \\
Serbian           & srp\_Cyrl           & $\xmark$ & $\cmark$  & Indo-European   & Balto-Slavic    & Cyrillic   & 5.5   & med  & 1.9    & low  \\
Silesian          & szl\_Latn           & $\xmark$ & $\xmark$ & Indo-European   & Balto-Slavic    & Latin      & 0.4   & low  & 0.0    & -    \\
Tatar             & tat\_Cyrl           & $\xmark$ & $\cmark$  & Turkic          & Common Turkic   & Cyrillic   & 2.1   & med  & 9.3    & -    \\
Tajik             & tgk\_Cyrl           & $\xmark$ & $\cmark$  & Indo-European   & Iranian         & Cyrillic   & 1.1   & med  & 0.0    & -    \\
Turkmen           & tuk\_Latn           & $\xmark$ & $\cmark$  & Turkic          & Common Turkic   & Latin      & 0.6   & low  & 0.8    & -    \\
Turkish           & tur\_Latn           & $\xmark$ & $\cmark$  & Turkic          & Common Turkic   & Latin      & 47.4  & high & 35.1   & high \\
Tuvan             & tyv\_Cyrl           & $\cmark$  & $\xmark$ & Turkic          & Common Turkic   & Cyrillic   & 0.2   & new  & 0.0    & -    \\
Uyghur            & uig\_Arab           & $\xmark$ & $\cmark$  & Turkic          & Common Turkic   & Arabic     & 0.8   & med  & 9.7    & -    \\
Ukrainian         & ukr\_Cyrl           & $\xmark$ & $\cmark$  & Indo-European   & Balto-Slavic    & Cyrillic   & 12.2  & med  & 25.1   & med  \\
Northern Uzbek    & uzn\_Latn           & $\xmark$ & $\cmark$  & Turkic          & Common Turkic   & Latin      & 4.1   & med  & 48.5   & high \\
Venetian          & vec\_Latn           & $\xmark$ & $\xmark$ & Indo-European   & Italic          & Latin      & 0.2   & low  & 0.0    & -    \\ \bottomrule
\end{tabular}%
        }
         \caption{Details about the languages used in our experiments.}
        \label{tab:data}
    \end{table*}

\section{Additional Results} \label{sec:appendix_results}

    In Table~\ref{tab:text_analytic_results} we present the per-language text translation results for the models of Table~\ref{tab:main_text_results}.
    
    In Table~\ref{tab:speech_analytic}, we present the per-language speech translation results for some of the models of Table~\ref{tab:main_speech_results}. We also add results from the SONAR Speech Encoders~\cite{sonar}, which were excluded from the main table since they do not support all the languages with which we experiment. Furthermore, Table~\ref{tab:speech_analytic} includes more than 33 languages, since for ease of presentation, in the main results we presented the ones that were both supported by Whisper and our models. We indicate the languages not taken into account for the results of Table~\ref{tab:main_speech_results}.

    In Tables~\ref{tab:res_text_per_category} and \ref{tab:new_res_text_per_category}, we present the per linguistic subgroup and script results of our text encoders in translation and cross-lingual similarity search. We observe that for known languages, the charSONAR encoder outperforms the subword-based encoder in all categories, apart from the single group that contains the Greek language, and only in cross-lingual similarity search. For the new languages, we notice that charSONAR performs better in all categories for translation, but the subword-based model is better for the Turkic and Uralic subgrouping, and Cyrillic script in cross-lingual similarity search.

    In Table~\ref{tab:group_encoders_dev} we present the text translation results for the three encoders that were used in the initial exploration with the four languages for the speech adaptation (Table~\ref{tab:speech_ablations}). We used the Uralic/Cyrillic encoder for Estonian and Russian, the Turkic for Turkish, and the Romance for Spanish.

\begin{figure}
        \centering
        \includegraphics[width=\columnwidth]{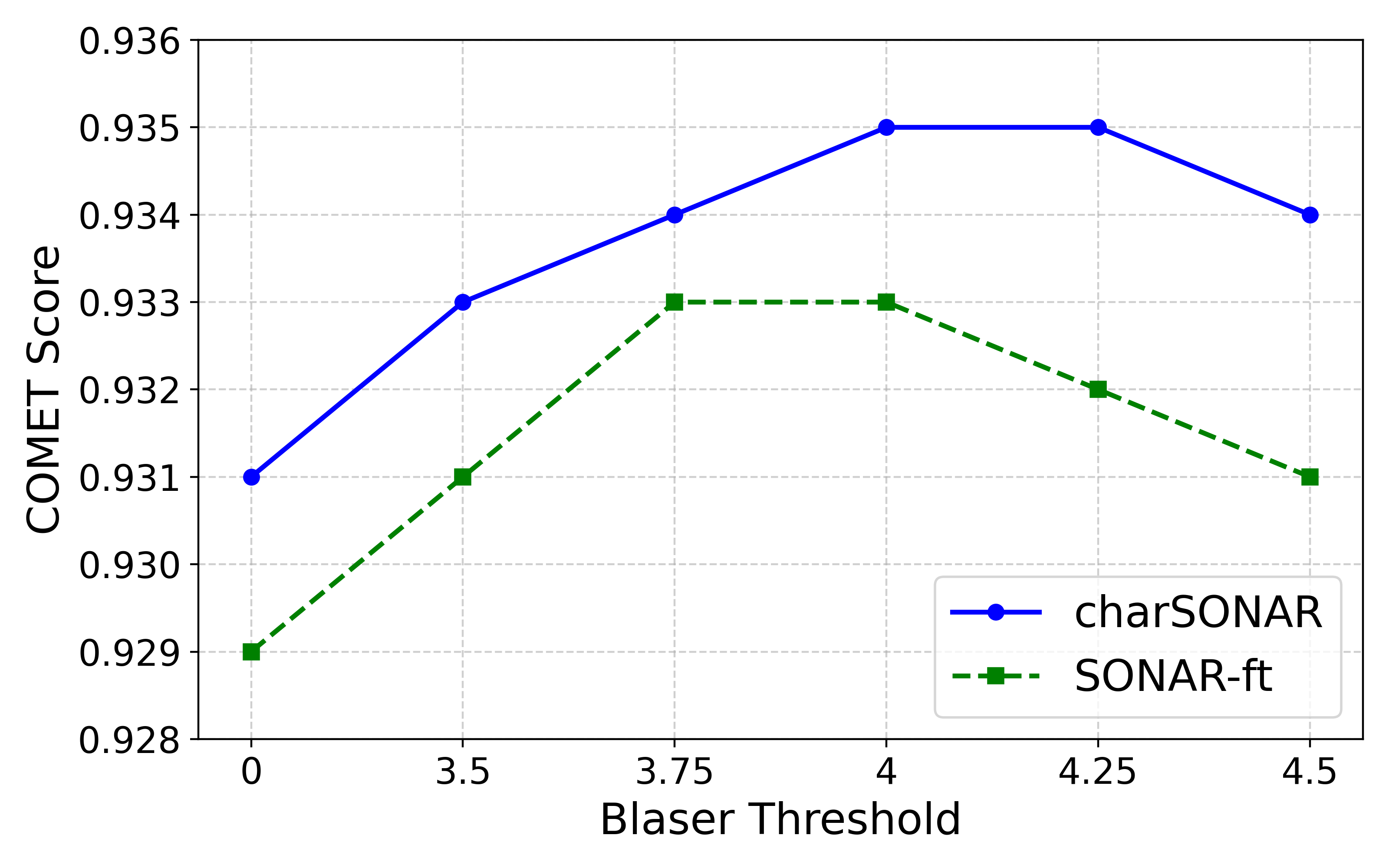}
         \caption{COMET scores vs. BLASER 2.0 filtering threshold for charSONAR and SONAR in FLORES \texttt{dev}. Results with the Ural/Cyrl group of 15 languages. COMET scores are average of X$\rightarrow$Eng for all the 15 languages in the group.}
        \label{fig:comet_vs_blaser_dev}
     \end{figure}

    In Figure~\ref{fig:comet_vs_blaser_dev} we present our ablation for deciding the BLASER 2.0 filtering threshold. To speed-up experimentation and use less data, we filtered with 4.5 for the initial exploration, but for the main experiments we used a threshold of 4.

    Finally, in Table~\ref{tab:efficiency_batching} we provide an efficiency analysis for SONAR and charSONAR models, similar to the results of \S\ref{subsec:efficiency_analysis}, but now with batching. We use length-based bucketing and a batch size of 5K tokens, which results in 8 batches for SONAR and 31 batches for charSONAR. The results here confirm the findings of Table~\ref{tab:efficiency}, showing that the impact of the char-based tokenization is minimal with respect to the additional computational overhead.

    \begin{table*}
    \centering
    \resizebox{0.73\textwidth}{!}{%
\begin{tabular}{@{}lcccccc@{}}
\toprule
\textbf{}           & \textbf{NLLB-200} & \textbf{SONAR-200} & \textbf{SONAR-63} & \textbf{charSONAR-63} & \textbf{SONAR-75} & \textbf{charSONAR-75} \\ \midrule
arg\_Latn           & 0.867$^\dagger$              & 0.860$^\dagger$               & 0.910$^\dagger$              & \underline{0.912}$^\dagger$            & \textbf{0.931}    & 0.917                 \\
ast\_Latn           & 0.918             & 0.909              & 0.920             & \textbf{0.928}        & 0.920             & 0.927                 \\
awa\_Deva           & \textbf{0.918}    & 0.861              & 0.894             & 0.896                 & 0.898             & 0.898                 \\
azb\_Arab           & 0.632             & 0.513              & \textbf{0.698}    & \textbf{0.698}        & 0.681             & 0.696                 \\
azj\_Latn           & \textbf{0.910}    & 0.741              & 0.863             & 0.868                 & 0.865             & 0.870                 \\
bak\_Cyrl           & 0.916             & 0.900              & 0.915             & \textbf{0.920}        & 0.916             & 0.918                 \\
bel\_Cyrl           & \textbf{0.928}    & 0.905              & 0.922             & 0.926                 & 0.920             & 0.926                 \\
bho\_Deva           & 0.907             & 0.879              & 0.902             & \textbf{0.908}        & 0.904             & 0.906                 \\
bos\_Latn           & \textbf{0.961}    & 0.953              & 0.960             & 0.960                 & 0.959             & 0.960                 \\
brx\_Deva$^*$              & 0.208$^\dagger$              & 0.203$^\dagger$               & \underline{0.217}$^\dagger$              & 0.216$^\dagger$                  & 0.841             & \textbf{0.851}        \\
bul\_Cyrl           & \textbf{0.954}    & 0.952              & 0.952             & 0.953                 & 0.952             & \textbf{0.954}        \\
cat\_Latn           & 0.954             & 0.949              & 0.952             & \textbf{0.956}        & 0.952             & \textbf{0.956}        \\
cat\_Latn\_vale1252 & 0.903$^\dagger$              & 0.886$^\dagger$               & 0.952$^\dagger$              & \underline{0.955}$^\dagger$            & 0.952$^\dagger$              & 0.955$^\dagger$                  \\
ces\_Latn           & 0.954             & 0.953              & 0.955             & \textbf{0.956}        & 0.955             & \textbf{0.956}        \\
chv\_Cyrl           & 0.240$^\dagger$              & 0.251$^\dagger$               & \underline{0.274}$^\dagger$        & 0.268$^\dagger$                  & 0.851             & \textbf{0.855}        \\
ckb\_Arab           & 0.881             & 0.872              & 0.882             & 0.889                 & 0.884             & \textbf{0.890}        \\
crh\_Latn           & 0.907             & 0.890              & 0.909             & \textbf{0.920}        & 0.914             & 0.918                 \\
dgo\_Deva$^*$           & \underline{0.606}$^\dagger$        & 0.567$^\dagger$               & 0.575$^\dagger$              & 0.590$^\dagger$                  & 0.895             & \textbf{0.899}        \\
ell\_Grek           & \textbf{0.940}    & 0.932              & 0.938             & 0.938                 & 0.937             & 0.939                 \\
est\_Latn           & 0.940             & 0.935              & 0.942             & \textbf{0.945}        & 0.942             & \textbf{0.945}        \\
fin\_Latn           & 0.942             & 0.936              & 0.945             & 0.946                 & 0.944             & \textbf{0.947}        \\
fra\_Latn           & \textbf{0.965}    & 0.961              & 0.961             & 0.961                 & 0.959             & 0.961                 \\
fur\_Latn           & 0.931             & 0.930              & 0.937             & \textbf{0.938}        & 0.936             & \textbf{0.938}        \\
glg\_Latn           & \textbf{0.957}    & 0.951              & 0.954             & \textbf{0.957}        & 0.954             & \textbf{0.957}        \\
gom\_Deva$^*$           & 0.511$^\dagger$              & 0.477$^\dagger$               & 0.607$^\dagger$              & \underline{0.635}$^\dagger$            & 0.869             & \textbf{0.880}        \\
hin\_Deva           & \textbf{0.939}    & 0.935              & 0.934             & 0.935                 & 0.934             & 0.938                 \\
hne\_Deva           & 0.911             & 0.902              & 0.913             & 0.915                 & 0.912             & \textbf{0.916}        \\
hrv\_Latn           & 0.947             & 0.949              & \textbf{0.954}    & \textbf{0.954}        & \textbf{0.954}    & \textbf{0.954}        \\
hun\_Latn           & 0.946             & 0.940              & 0.945             & \textbf{0.948}        & 0.945             & 0.947                 \\
ita\_Latn           & \textbf{0.957}    & 0.954              & 0.951             & 0.954                 & 0.952             & 0.954                 \\
kaa\_Latn           & 0.349$^\dagger$              & 0.359$^\dagger$              & 0.698$^\dagger$              & \underline{0.767}$^\dagger$            & 0.917             & \textbf{0.924}        \\
kas\_Deva           & 0.657             & 0.600              & 0.674             & 0.704                 & 0.681             & \textbf{0.710}        \\
kaz\_Cyrl           & \textbf{0.918}    & 0.908              & 0.912             & 0.917                 & 0.909             & 0.917                 \\
khk\_Cyrl           & 0.852             & 0.838              & 0.863             & 0.873                 & 0.865             & \textbf{0.876}        \\
kir\_Cyrl           & 0.908             & 0.898              & 0.911             & \textbf{0.915}        & 0.909             & \textbf{0.915}        \\
kmr\_Latn           & 0.784             & 0.776              & 0.789             & \textbf{0.804}        & 0.789             & 0.800                 \\
lij\_Latn           & 0.907             & 0.896              & 0.912             & \textbf{0.914}        & 0.910             & \textbf{0.914}        \\
lit\_Latn           & 0.930             & 0.921              & 0.933             & \textbf{0.938}        & 0.934             & 0.937                 \\
lmo\_Latn           & 0.866             & 0.833              & 0.884             & \textbf{0.900}        & 0.885             & 0.898                 \\
ltg\_Latn           & 0.888             & 0.866              & 0.900             & \textbf{0.917}        & 0.900             & 0.916                 \\
lvs\_Latn           & 0.928             & 0.915              & 0.932             & \textbf{0.938}        & 0.932             & 0.937                 \\
mag\_Deva           & 0.931             & 0.927              & 0.926             & \textbf{0.935}        & 0.929             & 0.930                 \\
mai\_Deva           & \textbf{0.930}    & 0.881              & 0.907             & 0.905                 & 0.906             & 0.906                 \\
mar\_Deva           & \textbf{0.929}    & 0.920              & 0.920             & 0.925                 & 0.921             & 0.927                 \\
mhr\_Cyrl$^*$        & 0.262$^\dagger$              & 0.278$^\dagger$               & 0.268$^\dagger$              & \underline{0.307}$^\dagger$            & 0.896             & \textbf{0.901}        \\
mkd\_Cyrl           & 0.946             & 0.942              & 0.946             & \textbf{0.952}        & 0.947             & 0.951                 \\
myv\_Cyrl           & 0.245$^\dagger$              & 0.243$^\dagger$               & 0.251$^\dagger$              & \underline{0.259}$^\dagger$            & \textbf{0.851}    & 0.846                 \\
npi\_Deva           & \textbf{0.926}    & 0.881              & 0.900             & 0.901                 & 0.900             & 0.901                 \\
oci\_Latn           & 0.956             & 0.952              & 0.958             & 0.958                 & 0.957             & \textbf{0.959}        \\
oci\_Latn\_aran1260 & 0.505$^\dagger$              & 0.500$^\dagger$               & 0.569$^\dagger$              & \underline{0.576}$^\dagger$            & 0.566$^\dagger$              & 0.571$^\dagger$                  \\
pbt\_Arab           & 0.866             & 0.855              & 0.870             & 0.872                 & 0.872             & \textbf{0.874}        \\
pes\_Arab           & 0.924             & 0.918              & 0.922             & \textbf{0.926}        & 0.925             & \textbf{0.926}        \\
pol\_Latn           & \textbf{0.950}    & 0.946              & 0.946             & 0.949                 & 0.948             & 0.948                 \\
por\_Latn           & \textbf{0.963}    & 0.960              & 0.961             & 0.962                 & 0.960             & 0.962                 \\
prs\_Arab           & 0.901             & 0.900              & 0.908             & \textbf{0.910}        & 0.908             & \textbf{0.910}        \\
ron\_Latn           & \textbf{0.964}    & 0.961              & 0.963             & \textbf{0.964}        & \textbf{0.964}    & \textbf{0.964}        \\
rus\_Cyrl           & \textbf{0.950}    & 0.942              & 0.944             & 0.946                 & 0.944             & 0.946                 \\
san\_Deva           & 0.749             & 0.702              & 0.731             & 0.737                 & 0.732             & 0.742                 \\
scn\_Latn           & 0.896             & 0.877              & 0.904             & \textbf{0.915}        & 0.905             & 0.914                 \\
slk\_Latn           & 0.955             & 0.951              & 0.953             & \textbf{0.956}        & 0.952             & 0.955                 \\
slv\_Latn           & 0.944             & 0.943              & 0.948             & \textbf{0.951}        & 0.948             & \textbf{0.951}        \\
snd\_Deva$^*$           & 0.466$^\dagger$             & 0.464$^\dagger$               & \underline{0.512}$^\dagger$              & 0.497$^\dagger$                  & 0.860             & \textbf{0.869}        \\
spa\_Latn           & \textbf{0.950}    & 0.949              & 0.946             & 0.948                 & 0.946             & 0.948                 \\
srd\_Latn           & 0.909             & 0.902              & 0.917             & \textbf{0.918}        & 0.915             & 0.916                 \\
srp\_Cyrl           & 0.949             & 0.943              & 0.950             & \textbf{0.954}        & 0.950             & \textbf{0.954}        \\
szl\_Latn           & 0.930             & 0.920              & 0.934             & \textbf{0.942}        & 0.933             & 0.940                 \\
tat\_Cyrl           & 0.927             & 0.915              & 0.925             & 0.927                 & 0.925             & \textbf{0.928}        \\
tgk\_Cyrl           & 0.915             & 0.903              & 0.915             & \textbf{0.925}        & 0.916             & 0.923                 \\
tuk\_Latn           & 0.893             & 0.882              & 0.898             & 0.910                 & 0.904             & \textbf{0.912}        \\
tur\_Latn           & \textbf{0.946}    & 0.940              & 0.943             & \textbf{0.946}        & 0.945             & 0.945                 \\
tyv\_Cyrl           & 0.281$^\dagger$              & 0.313$^\dagger$               & 0.366$^\dagger$              & \underline{0.381}$^\dagger$            & 0.880             & \textbf{0.884}        \\
uig\_Arab           & 0.863             & 0.853              & 0.859             & \textbf{0.868}        & 0.859             & 0.867                 \\
ukr\_Cyrl           & 0.948             & 0.942              & 0.945             & 0.948                 & 0.946             & \textbf{0.949}        \\
uzn\_Latn           & \textbf{0.931}    & 0.914              & 0.918             & 0.924                 & 0.920             & 0.926                 \\
vec\_Latn           & 0.920             & 0.907              & 0.931             & 0.933                 & 0.930             & \textbf{0.936}        \\ \bottomrule
\end{tabular}%
    }
     \caption{Text Translation COMET (X$\rightarrow$Eng) scores in FLORES \texttt{devtest}. $*$ indicates translation is evaluated on \texttt{dev} split. $\dagger$ indicates that the result is zero-shot. \underline{Underlined} is the best among the zero-shot for each language, if any. In \textbf{bold} is the best for supervised results for each language, if any.}
    \label{tab:text_analytic_results}
    \end{table*}
    
    \begin{table*}
    \centering
    \resizebox{\textwidth}{!}{%
\begin{tabular}{@{}lcccccccccc@{}}
\toprule
                  & \multicolumn{3}{c}{\textbf{E2E ST}}                      & \multicolumn{2}{c}{\textbf{charSONAR Cascades}} & \multicolumn{3}{c}{\textbf{Speech-charSONAR (CV)}}     & \multicolumn{2}{c}{\textbf{+ FLEURS}} \\ \midrule
\textbf{Language} & \textbf{SONAR} & \textbf{Whisper} & \textbf{SeamlessM4T} & \textbf{w/ MMS}      & \textbf{w/ Whisper}      & \textbf{Random} & \textbf{Pretrained} & \textbf{Dual}  & \textbf{Pretrained}  & \textbf{Dual}  \\ \midrule
ast\_Latn$^*$     & -              & -                & 0.752                & 0.821                & -                        & 0.204           & 0.816               & 0.223          & 0.840                & \textbf{0.849} \\
azj\_Latn         & -              & 0.642            & \textbf{0.837}       & 0.762                & 0.777                    & 0.197           & 0.788               & 0.194          & 0.788                & 0.764          \\
bel\_Cyrl         & 0.823          & 0.676            & 0.885                & 0.864                & 0.842                    & 0.856           & 0.864               & 0.868          & 0.874                & \textbf{0.886} \\
bos\_Latn         & 0.882          & 0.829            & \textbf{0.919}       & 0.899                & 0.911                    & -               & -                   & -              & -                    & -              \\
bul\_Cyrl         & 0.852          & 0.816            & 0.902                & 0.884                & 0.891                    & 0.879           & 0.891               & 0.898          & 0.892                & \textbf{0.907} \\
cat\_Latn         & 0.886          & 0.874            & 0.929                & 0.893                & 0.929                    & 0.903           & 0.900               & 0.922          & 0.913                & \textbf{0.930} \\
ces\_Latn         & 0.873          & 0.811            & 0.910                & 0.889                & 0.904                    & 0.888           & 0.890               & 0.904          & 0.902                & \textbf{0.911} \\
ckb\_Arab$^*$     & -              & -                & 0.766                & 0.786                & -                        & 0.692           & 0.800               & 0.798          & 0.802                & \textbf{0.818} \\
ell\_Grek         & -              & 0.747            & 0.868                & 0.860                & 0.874                    & 0.216           & 0.866               & 0.866          & 0.872                & \textbf{0.889} \\
est\_Latn         & 0.805          & 0.629            & 0.898                & 0.908                & 0.902                    & 0.877           & 0.910               & 0.912          & \textbf{0.918}       & 0.914          \\
fin\_Latn         & 0.799          & 0.768            & 0.887                & 0.900                & 0.926                    & 0.229           & 0.902               & 0.904          & \textbf{0.906}       & 0.905          \\
fra\_Latn         & 0.870          & 0.891            & 0.910                & 0.884                & \textbf{0.935}           & 0.888           & 0.906               & 0.914          & 0.914                & 0.922          \\
glg\_Latn         & -              & 0.830            & 0.917                & 0.897                & 0.908                    & 0.859           & 0.908               & \textbf{0.913} & 0.906                & 0.910          \\
hin\_Deva         & 0.770          & 0.746            & 0.856                & 0.856                & 0.834                    & 0.854           & 0.876               & 0.875          & 0.880                & \textbf{0.882} \\
hrv\_Latn$^*$     & 0.866          & 0.816            & 0.895                & 0.903                & \textbf{0.912}           & -               & -                   & -              & -                    & -              \\
hun\_Latn         & -              & 0.724            & 0.878                & 0.867                & 0.886                    & 0.870           & 0.866               & 0.882          & 0.884                & \textbf{0.892} \\
ita\_Latn         & 0.892          & 0.898            & 0.927                & 0.915                & \textbf{0.943}           & 0.914           & 0.924               & 0.926          & 0.932                & 0.939          \\
kaz\_Cyrl         & -              & 0.349            & 0.846                & 0.844                & 0.774                    & 0.198           & 0.849               & 0.198          & 0.866                & \textbf{0.876} \\
khk\_Cyrl         & -              & 0.207            & 0.748                & 0.731                & 0.342                    & 0.229           & 0.732               & 0.213          & 0.763                & \textbf{0.764} \\
kir\_Cyrl$^*$     & -              & -                & 0.854                & 0.837                & -                        & 0.197           & 0.855               & 0.853          & 0.861                & \textbf{0.865} \\
lit\_Latn         & 0.766          & 0.579            & 0.832                & 0.880                & 0.853                    & 0.858           & 0.881               & 0.885          & 0.884                & \textbf{0.897} \\
lvs\_Latn         & 0.848          & 0.589            & 0.885                & 0.899                & 0.888                    & 0.879           & 0.905               & 0.909          & 0.908                & \textbf{0.910} \\
mar\_Deva         & 0.734          & 0.503            & 0.821                & 0.812                & 0.684                    & 0.734           & 0.836               & 0.843          & 0.840                & \textbf{0.846} \\
mkd\_Cyrl         & 0.887          & 0.808            & 0.917                & 0.919                & 0.912                    & 0.246           & 0.927               & 0.925          & \textbf{0.928}       & \textbf{0.928} \\
npi\_Deva         & 0.675          & 0.538            & \textbf{0.826}       & 0.790                & 0.652                    & 0.196           & 0.810               & 0.798          & 0.801                & 0.799          \\
oci\_Latn         & -              & 0.483            & 0.568                & 0.747                & 0.583                    & 0.202           & 0.707               & 0.707          & 0.795                & \textbf{0.805} \\
pes\_Arab         & 0.810          & 0.666            & 0.887                & 0.867                & 0.851                    & 0.875           & 0.890               & 0.884          & 0.887                & \textbf{0.899} \\
pol\_Latn         & 0.860          & 0.856            & 0.893                & 0.888                & \textbf{0.923}           & 0.876           & 0.888               & 0.899          & 0.898                & 0.908          \\
por\_Latn         & 0.878          & 0.906            & 0.897                & 0.902                & \textbf{0.941}           & 0.885           & 0.908               & 0.917          & 0.918                & 0.922          \\
ron\_Latn         & 0.856          & 0.867            & 0.909                & 0.895                & \textbf{0.919}           & 0.855           & 0.900               & 0.904          & 0.906                & 0.905          \\
rus\_Cyrl         & 0.878          & 0.893            & \textbf{0.912}       & 0.883                & 0.934                    & 0.883           & 0.898               & 0.908          & 0.903                & 0.910          \\
slk\_Latn         & 0.885          & 0.822            & 0.914                & 0.911                & 0.924                    & 0.876           & 0.909               & 0.917          & 0.919                & \textbf{0.921} \\
slv\_Latn         & 0.843          & 0.672            & 0.879                & 0.871                & 0.852                    & 0.201           & 0.871               & 0.200          & \textbf{0.883}       & 0.875          \\
snd\_Deva$^*$     & 0.360          & 0.360            & 0.443                & \textbf{0.698}       & 0.423                    & -               & -                   & -              & -                    & -              \\
spa\_Latn         & 0.888          & 0.893            & 0.908                & 0.899                & \textbf{0.934}           & 0.912           & 0.917               & 0.920          & 0.922                & 0.930          \\
srp\_Cyrl         & 0.891          & 0.856            & 0.924                & 0.924                & \textbf{0.929}           & 0.235           & 0.922               & 0.913          & 0.927                & 0.928          \\
tgk\_Cyrl$^*$     & -              & 0.523            & 0.858                & \textbf{0.874}       & 0.658                    & -               & -                   & -              & -                    & -              \\
tur\_Latn         & 0.743          & 0.827            & 0.888                & 0.878                & \textbf{0.924}           & 0.868           & 0.885               & 0.903          & 0.899                & 0.909          \\
ukr\_Cyrl         & 0.858          & 0.865            & 0.912                & 0.890                & \textbf{0.931}           & 0.887           & 0.903               & 0.904          & 0.911                & 0.915          \\
uzn\_Latn         & 0.736          & 0.326            & 0.846                & 0.753                & 0.527                    & 0.798           & 0.839               & 0.843          & 0.846                & \textbf{0.854} \\ \bottomrule
\end{tabular}%
    }
     \caption{Speech Translation COMET scores (X$\rightarrow$Eng) on FLEURS \texttt{test}. The 6 languages with $*$ where not part of the main results of Table~\ref{tab:main_speech_results}, since they were not supported either by our models or by Whisper.}
    \label{tab:speech_analytic}
    \end{table*}

    \begin{table*}[ht]
        \centering
        \resizebox{0.7\textwidth}{!}{%
        \begin{tabular}{lcccc|ccc}
        \toprule
        & \multicolumn{4}{c}{\textbf{Subgrouping}} & \multicolumn{3}{c}{\textbf{Script}} \\
        \cmidrule(lr){2-5} \cmidrule(lr){6-8}
        \textbf{} & \textbf{Indic} & \textbf{Romance} & \textbf{Turkic} & \textbf{Uralic} & \textbf{Devanagari} & \textbf{Latin} & \textbf{Cyrillic} \\
        \midrule
        \textbf{$\#$ Languages} & 4 & 3 & 3 & 2 & 4 & 4 & 4 \\
        \midrule
        \textbf{COMET} & & & & & & & \\
        SONAR & 0.428 & 0.749 & 0.305 & 0.260 & 0.428 & 0.651 & 0.269 \\
        SONAR-75 & 0.866 & 0.813 & 0.882 & 0.869 & 0.866 & 0.839 & 0.867 \\
        charSONAR-75 & \textbf{0.875} & \textbf{0.814} & \textbf{0.888} & \textbf{0.874} & \textbf{0.875} & \textbf{0.842} & \textbf{0.871} \\
        \midrule
        \textbf{XSIM++} & & & & & & & \\
        SONAR-200 & 53.8 & 29.3 & 63.5 & 70.2 & 53.8 & 36.1 & 68.6 \\
        SONAR-75 & 8.7 & 21.6 & \textbf{9.6} & \textbf{10.5} & 8.7 & 17.7 & \textbf{11.0} \\
        charSONAR-75 & \textbf{8.2} & \textbf{20.9} & 10.1 & 10.8 & \textbf{8.2} & \textbf{16.9} & 11.8 \\
        \bottomrule
        \end{tabular}%
        }
         \caption{Text translation (COMET) and text retrieval (xSIM++) results per language subgroup and script for the 12 newly added languages. Results in FLORES \texttt{devtest} (X$\rightarrow$Eng).}
        \label{tab:new_res_text_per_category}
    \end{table*}
    
    \begin{table*}[ht]
        \centering
        \setlength{\tabcolsep}{5pt} 
        \resizebox{\textwidth}{!}{%
        \begin{tabular}{lcccccccc|ccccc}
        \toprule
        & \multicolumn{8}{c}{\textbf{Subgrouping}} & \multicolumn{5}{c}{\textbf{Script}} \\
        \cmidrule(lr){2-9} \cmidrule(lr){10-14}
        \textbf{} & \textbf{Balto-Slavic} & \textbf{Romance} & \textbf{Turkic} & \textbf{Indic} & \textbf{Iranian} & \textbf{Uralic} & \textbf{Mongolic} & \textbf{Greek} & \textbf{Latin} & \textbf{Cyrillic} & \textbf{Devanagari} & \textbf{Arabic} & \textbf{Greek} \\
        \midrule
        \textbf{$\#$ Languages} & 16 & 15 & 11 & 10 & 6 & 3 & 1 & 1 & 34 & 12 & 10 & 6 & 1 \\
        \midrule
        \textbf{COMET} & & & & & & & & & & & & & \\
        SONAR-200 & 0.934 & 0.926 & 0.850 & 0.849 & 0.871 & 0.937 & 0.838 & 0.932 & 0.917 & 0.916 & 0.849 & 0.818 & 0.932 \\
        SONAR-75 & 0.942 & 0.935 & 0.884 & 0.870 & 0.881 & 0.944 & 0.864 & 0.936 & 0.929 & 0.925 & 0.870 & 0.852 & 0.936 \\
        charSONAR-75 & \textbf{0.946} & \textbf{0.940} & \textbf{0.892} & \textbf{0.877} & \textbf{0.887} & \textbf{0.946} & \textbf{0.876} & \textbf{0.939} & \textbf{0.934} & \textbf{0.930} & \textbf{0.877} & \textbf{0.860} & \textbf{0.939} \\
        \midrule
        \textbf{XSIM++} & & & & & & & & & & & & & \\
        SONAR & 8.1 & 8.0 & 13.4 & 13.5 & 10.7 & 7.1 & 13.3 & 9.2 & 8.3 & 9.7 & 13.5 & 16.1 & 9.2 \\
        SONAR-75 & 6.2 & 5.5 & 10.0 & 9.7 & 8.2 & 5.7 & 11.0 & \textbf{6.7} & 6.1 & 7.7 & 9.7 & 11.6 & \textbf{6.7} \\
        charSONAR & \textbf{5.8} & \textbf{5.2} & \textbf{9.4} & \textbf{9.1} & \textbf{7.7} & \textbf{5.3} & \textbf{10.1} & 6.9 & \textbf{5.7} & \textbf{7.1} & \textbf{9.1} & \textbf{11.0} & 6.9 \\
        \bottomrule
        \end{tabular}%
        }
         \caption{Text translation (COMET) and text retrieval (xSIM++) results per language subgroup and script for the 63 known training languages. Results in FLORES \texttt{devtest} (X$\rightarrow$Eng).}
        \label{tab:res_text_per_category}
    \end{table*}
    
    \begin{table}
    \centering
    \resizebox{\columnwidth}{!}{%
    \begin{tabular}{@{}lcccc@{}}
    \toprule
    \textbf{Model}                     & \textbf{Ural/Cyrl} & \textbf{Turkic} & \textbf{Romance} & \textbf{Avg}   \\ \midrule
    SONAR-200   & 0.925 & 0.857 & 0.932 & 0.905 \\ \midrule
    SONAR-group & 0.930 & 0.879 & 0.942 & 0.917 \\
    charSONAR-group                    & \textbf{0.934}     & \textbf{0.881}  & 0.946            & \textbf{0.920} \\
    $\hookrightarrow$ w/ Norm          & \textbf{0.934}     & 0.877           & 0.946            & 0.919          \\
    $\hookrightarrow$ w/ Norm \& Noise & \textbf{0.934}     & 0.878           & \textbf{0.947}   & \textbf{0.920} \\ \bottomrule
    \end{tabular}%
    }
     \caption{Text Translation COMET scores (X$\rightarrow$Eng) in FLORES \texttt{dev}. Each encoder was trained on the respective group of languages.}
    \label{tab:group_encoders_dev}
     \end{table}

    \begin{table}
        \centering
        \resizebox{0.75\columnwidth}{!}{%
        \begin{tabular}{@{}lcc@{}}
        \toprule
        \textbf{Model} & \textbf{\# Tokens} & \textbf{\begin{tabular}[c]{@{}c@{}}Inference\\ Time (s)\end{tabular}} \\ \midrule
        SONAR          & 49\phantom{ ($\times$3.2)}                & 127\phantom{ ($\times$1.1)}                                                              \\
        charSONAR      & 158 ($\times$3.2) & 142 (+$10\%$)                                               \\ \bottomrule
        \end{tabular}
         }
        \caption{Average number of tokens and average inference time in FLORES \texttt{dev} with batching (5K tokens per batch).}
        \label{tab:efficiency_batching}
    \end{table}


\end{document}